\documentclass{article}

\usepackage[accepted]{icml2024}

\usepackage{amsmath, amssymb, amsfonts, mathtools}
\usepackage{physics}
\usepackage{cancel}
\usepackage{thmtools, thm-restate}
\usepackage{accents}
\usepackage{bm}

\DeclareMathOperator{\adj}{\mathsf{Adj}}

\DeclareMathOperator{\argmin}{arg min}

\makeatletter
\newcommand{\ostar}{\mathbin{\mathpalette\make@circled *}}
\newcommand{\make@circled}[2]{%
  \ooalign{$\m@th#1\smallbigcirc{#1}$\cr\hidewidth$\m@th#1#2$\hidewidth\cr}%
}
\newcommand{\smallbigcirc}[1]{%
  \vcenter{\hbox{\scalebox{0.77778}{$\m@th#1\bigcirc$}}}%
}
\makeatother

\newcommand{\x}{\times}

\usepackage{pict2e, picture}
\makeatletter
\DeclareRobustCommand{\Arrow}[1][]{%
\check@mathfonts
\if\relax\detokenize{#1}\relax
\settowidth{\dimen@}{$\m@th\rightarrow$}%
\else
\setlength{\dimen@}{#1}%
\fi
\sbox\z@{\usefont{U}{lasy}{m}{n}\symbol{41}}%
\begin{picture}(\dimen@,\ht\z@)
\roundcap
\put(\dimexpr\dimen@-.7\wd\z@,0){\usebox\z@}
\put(0,\fontdimen22\textfont2){\line(1,0){\dimen@}}
\end{picture}%
}
\makeatother

\usepackage{scalerel}
\newcommand{\cH}{\mathcal{H}}

\newcommand{\cO}{\mathcal{O}}

\newcommand{\cF}{\mathcal{F}}

\newcommand{\cZ}{\mathcal{Z}}
 
\newcommand{\sA}{\mathsf{A}}
\newcommand{\sB}{\mathsf{B}}
\newcommand{\sC}{\mathsf{C}}
\newcommand{\sD}{\mathsf{D}}

\newcommand{\sG}{\mathsf{G}}

\newcommand{\sI}{\mathsf{I}}

\newcommand{\sK}{\mathsf{K}}
\newcommand{\sL}{\mathsf{L}}
\newcommand{\sM}{\mathsf{M}}

\newcommand{\sT}{\mathsf{T}}

\newcommand{\bC}{\mathbb{C}}
\newcommand{\bD}{\mathbb{D}}

\newcommand{\bN}{\mathbb{N}}

\newcommand{\R}{\mathbb{R}}
\newcommand{\bS}{\mathbb{S}}
\newcommand{\bT}{\mathbb{T}}

\DeclareMathAlphabet{\nummathbb}{U}{BOONDOX-ds}{m}{n}

\newcommand{\w}{\omega}

\usepackage{tikz}

\usepackage{pifont}
\newcommand{\xmark}{\ding{55}}%

\definecolor{amber}{rgb}{1.0, 0.49, 0.0}
\definecolor{amethyst}{rgb}{0.6, 0.4, 0.8}
\definecolor{asparagus}{rgb}{0.53, 0.66, 0.42}
\definecolor{floralwhite}{rgb}{1.0, 0.98, 0.94}
\definecolor{ghostwhite}{rgb}{0.97, 0.97, 1.0}
\definecolor{americanrose}{rgb}{1.0, 0.01, 0.24}
\definecolor{aqua}{rgb}{0.0, 1.0, 1.0}
\definecolor{electricpurple}{rgb}{0.75, 0.0, 1.0}
\definecolor{electricultramarine}{rgb}{0.25, 0.0, 1.0}
\definecolor{palatinateblue}{rgb}{0.15, 0.23, 0.89}
\definecolor{persiangreen}{rgb}{0.0, 0.65, 0.58}
\definecolor{turquoiseblue}{rgb}{0.0, 1.0, 0.94}
\definecolor{turquoise}{rgb}{0.19, 0.84, 0.78}
\definecolor{blue-green}{rgb}{0.0, 0.87, 0.87}
\definecolor{deepcarminepink}{rgb}{0.94, 0.19, 0.22}
\definecolor{vividcerise}{rgb}{0.85, 0.11, 0.51}
\definecolor{rose}{rgb}{1.0, 0.0, 0.5}
\definecolor{zaffre}{rgb}{0.0, 0.08, 0.66}
\definecolor{warmblack}{rgb}{0.0, 0.26, 0.26}
\definecolor{winter0.0}{rgb}{0.0, 0.0, 1.0}
\definecolor{winter0.2}{rgb}{0.0, 0.2, 0.9}
\definecolor{winter0.4}{rgb}{0.0, 0.4, 0.8}
\definecolor{winter0.6}{rgb}{0.0, 0.6, 0.7}
\definecolor{winter0.8}{rgb}{0.0, 0.8, 0.6}
\definecolor{winter1.0}{rgb}{0.0, 1.0, 0.5}

\definecolor{mayablue}{rgb}{0.45, 0.76, 0.98}

\definecolor{highlighteq}{rgb}{0.44, 0.16, 0.39}
\usetikzlibrary{patterns,snakes,positioning} 

\tikzset{
params/.style = {rectangle, draw=black, minimum size=5mm},
func/.style = {rectangle, draw=black, minimum size=5mm, rounded corners},
pad/.style = {rectangle, draw=winter0.2, fill=white, ultra thick, minimum size=10mm},
freq/.style = {rectangle, draw=winter0.4, fill=winter0.4!20, ultra thick, minimum size=10mm},
kernel/.style = {rectangle, draw=winter0.6, fill=winter0.6!20, ultra thick, minimum size=10mm},
}

\usepackage{booktabs} 
\usepackage{caption}
\usepackage{subcaption} 
\usepackage[bookmarks=true, 
            colorlinks=true,
            linkcolor=magenta,
            citecolor=blue!70,
            ]{hyperref}
\usepackage[textsize=tiny]{todonotes}
\usepackage{listings}
\usepackage{xcolor} 
\usepackage{diagbox} 
\usepackage{amsthm}
\usepackage{multirow}
\usepackage[many]{tcolorbox}
\tcbuselibrary{breakable,xparse,skins}
\definecolor{codegreen}{rgb}{0,0.6,0}
\definecolor{codegray}{rgb}{0.5,0.5,0.5}
\definecolor{codepurple}{rgb}{0.58,0,0.82}
\definecolor{backcolour}{rgb}{0.95,0.95,0.92}
\definecolor{ghostwhite}{rgb}{0.97, 0.97, 1.0}

\lstdefinestyle{mystyle}{
    backgroundcolor=\color{ghostwhite},   
    commentstyle=\color{codegreen},
    keywordstyle=\color{magenta},
    numberstyle=\tiny\color{codegray},
    stringstyle=\color{codepurple},
    basicstyle=\ttfamily\footnotesize,
    breakatwhitespace=false,         
    breaklines=true,                 
    captionpos=b,                    
    keepspaces=true,                 
    numbers=left,                    
    numbersep=5pt,                  
    showspaces=false,                
    showstringspaces=false,
    showtabs=false,                  
    tabsize=2
}

\usepackage[shortlabels]{enumitem}
\usepackage{authblk}
\usepackage{multicol}

\theoremstyle{plain}
\newtheorem{theorem}{Theorem}[section]

\newtheorem{lemma}[theorem]{Lemma}

\theoremstyle{definition}

\theoremstyle{remark}

\icmltitlerunning{State-Free Inference of State-Space Models}

\begin{document}
\setlength{\parskip}{4pt}
\twocolumn[
\icmltitle{State-Free Inference of State-Space Models: \\
The \textit{Transfer Function} Approach}

\icmlsetsymbol{equal}{*}
\icmlsetsymbol{sequal}{$\dagger$}

\begin{icmlauthorlist}
\icmlauthor{Rom N. Parnichkun}{liquid,tokyo,equal}
\icmlauthor{Stefano Massaroli}{liquid,riken,equal} 
\icmlauthor{Alessandro Moro}{tokyo,equal} \\
\icmlauthor{Jimmy T.H. Smith}{liquid,st}
\icmlauthor{Ramin Hasani}{liquid,mit}
\icmlauthor{Mathias Lechner}{liquid,mit} 
\icmlauthor{Qi An}{tokyo} \\
\icmlauthor{Christopher Ré}{st}
\icmlauthor{Hajime Asama}{tokyo}
\icmlauthor{Stefano Ermon}{st}
\icmlauthor{Taiji Suzuki}{tokyo,riken} \\
\icmlauthor{Atsushi Yamashita}{tokyo,sequal}
\icmlauthor{Michael Poli}{liquid,st,sequal}

\end{icmlauthorlist}

\icmlaffiliation{tokyo}{The University of Tokyo}
\icmlaffiliation{st}{Stanford University}
\icmlaffiliation{liquid}{Liquid AI}
\icmlaffiliation{mit}{Massachusetts Institute of Technology}
\icmlaffiliation{riken}{RIKEN}

\icmlcorrespondingauthor{Rom N. Parnichkun}{parnichkun@robot.t.u-tokyo.ac.jp}

\icmlkeywords{State-space model, transfer function, signal processing, control theory, language modeling}

\vskip 0.3in

]

\printAffiliationsAndNotice{$^*$Equal contribution $^\dagger$Equal senior authorship}

\begin{abstract}
We approach designing a state-space model for deep learning applications through its dual representation, the \textit{transfer function}, and uncover a highly efficient sequence parallel inference algorithm that is \textit{state-free}: unlike other proposed algorithms, state-free inference does not incur any significant memory or computational cost with an increase in state size. 
We achieve this using properties of the proposed frequency domain transfer function parametrization, which enables direct computation of its corresponding convolutional kernel's spectrum via a single Fast Fourier Transform. Our experimental results across multiple sequence lengths and state sizes illustrates, on average, a 35\% training speed improvement over S4 layers -- parametrized in time-domain -- on the Long Range Arena benchmark, while delivering state-of-the-art downstream performances over other attention-free approaches. Moreover, we report improved perplexity in language modeling over a long convolutional Hyena baseline, by simply introducing our transfer function parametrization. Our code is available at \url{https://github.com/ruke1ire/RTF}.

\end{abstract}

 \section{Introduction}

Central to the success of a certain class of sequence modeling layers are linear recurrences, which unlike the nonlinear case \cite{lstm, gru,kidger2020neural,massaroli2021differentiable}, are compatible with exact sequence parallel algorithms i.e., parallel scans \cite{sum_prefix,parallelrnn, s5, mamba, gateloop}, or (with time-invariance) the Fast Fourier Transform (FFT) \cite{s4, s4d, spacetime}. Such recurrent layers, often referred to in deep learning simply as \textit{state-space models}, depending on their parametrization, also boast efficient constant time and memory autoregressive inference, lowering latency and memory costs. 

Despite recent advancements, current SSMs exhibit certain limitations that this paper aims to address.





\begin{figure}
    \centering
    \includegraphics[trim={0 3mm 0 0},clip, width=.45\textwidth]{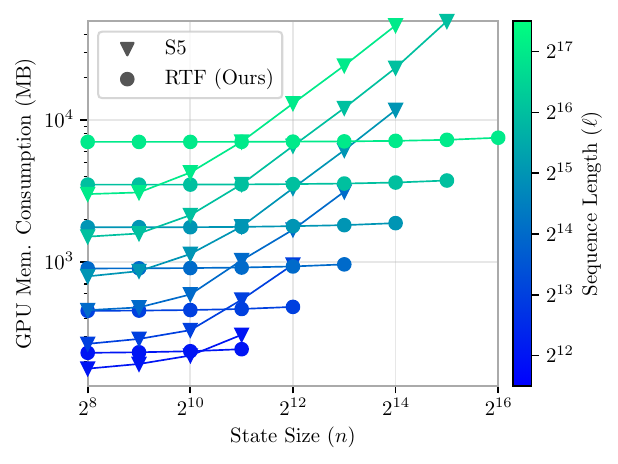}
    \caption{An illustration depicting the scaling of memory consumption on a scan-based algorithm (S5) and the proposed state-free inference algorithm denoted as RTF. We note that with larger state sizes, inference with S5 becomes prohibitively memory-intensive.}
    \label{fig:mem}
\end{figure}

\definecolor{amber}{rgb}{1.0, 0.49, 0.0}
\definecolor{h}{rgb}{0.0, 0.08, 0.66}
\definecolor{b}{rgb}{0.6, 0.4, 0.8}
\definecolor{a}{rgb}{0.85, 0.11, 0.51}
\definecolor{asparagus}{rgb}{0.53, 0.66, 0.42}
\definecolor{floralwhite}{rgb}{1.0, 0.98, 0.94}
\definecolor{ghostwhite}{rgb}{0.97, 0.97, 1.0}
\definecolor{americanrose}{rgb}{1.0, 0.01, 0.24}
\definecolor{aqua}{rgb}{0.0, 1.0, 1.0}
\definecolor{electricpurple}{rgb}{0.75, 0.0, 1.0}
\definecolor{electricultramarine}{rgb}{0.25, 0.0, 1.0}
\definecolor{palatinateblue}{rgb}{0.15, 0.23, 0.89}
\definecolor{persiangreen}{rgb}{0.0, 0.65, 0.58}
\definecolor{turquoiseblue}{rgb}{0.0, 1.0, 0.94}
\definecolor{turquoise}{rgb}{0.19, 0.84, 0.78}
\definecolor{blue-green}{rgb}{0.0, 0.87, 0.87}
\definecolor{deepcarminepink}{rgb}{0.94, 0.19, 0.22}
\definecolor{rose}{rgb}{1.0, 0.0, 0.5}
\definecolor{warmblack}{rgb}{0.0, 0.26, 0.26}
\definecolor{winter0.0}{rgb}{0.0, 0.0, 1.0}
\definecolor{winter0.2}{rgb}{0.0, 0.2, 0.9}
\definecolor{winter0.4}{rgb}{0.0, 0.4, 0.8}
\definecolor{winter0.6}{rgb}{0.0, 0.6, 0.7}
\definecolor{winter0.8}{rgb}{0.0, 0.8, 0.6}
\definecolor{winter1.0}{rgb}{0.0, 1.0, 0.5}

\begin{figure*}
\centering
\begin{tikzpicture}
\node[draw=black!50, rounded corners, fill=ghostwhite, text width=\textwidth, minimum height=70mm] (bg) at (0,0) {};
\node[text width=\textwidth, align=center] (top_heading) [below=2mm of bg.north] {\textit{\textbf{Rational Transfer Function (a)}} \vspace{2mm} \\ { $H(z) = {\color{h}h_0} + \dfrac{{\color{b}b_1}z^{-1} + \cdots + {\color{b}b_n} z^{-n}}{1 + {\color{a}a_1}z^{-1} + \cdots + {\color{a}a_n}z^{-n}} ={\color{h}h_0} + \dfrac{0 + {\color{b}b_1}z^{-1} + \cdots + {\color{b}b_n} z^{-n} + 0z^{-n-1} + \cdots + 0z^{-\ell+1}}{1 + {\color{a}a_1}z^{-1} + \cdots + {\color{a}a_n}z^{-n} + 0z^{-n-1} + \cdots + 0z^{-\ell +1}} $}};

\node[minimum height=55mm, text width=\textwidth] (bg_inf) [below=0mm of top_heading.south] {};
\node[text width=.5\textwidth, align=center] (par_inf_heading) [below=2mm of bg_inf.north west, anchor=north west] {\textit{\textbf{State-Free Parallel Inference (b)}}};

\node[params, draw=b, text width=25mm, align=center, anchor=north west] (bs) [below right=4mm and 4mm of par_inf_heading.south west] {\small $0\; {\color{b}b_1 \cdots b_n} \; 0 \; \cdots \; 0$};
\node[params, draw=a, text width=25mm, align=center] (as) [below=1mm of bs] {\small $1\; {\color{a}a_1  \cdots  a_n} \; 0 \; \cdots \; 0$};
\node[func, align=center, minimum height=12mm] (rfft_1) [below right=-5.5mm and 4mm of bs] {\small \tt rFFT};
\node[func, align=center, minimum height=12mm, text width=10mm] (rational_func) [right=4mm of rfft_1] {\small $\dfrac{\color{b}\mathbf{b}}{\color{a}\mathbf{a}}+{\color{h}h_0}$};
\node[func, align=center, minimum height=6mm] (irfft_1) [right=4mm of rational_func] {\small \tt irFFT};
\node[params, text width=15mm, align=center] (hs) [below=10mm of irfft_1.center] {\small ${\color{h} h_0} \cdots h_{\ell-1}$};
\node[params, text width=15mm, align=center] (us) [below=1mm of hs] {\small $u_0 \cdots u_{\ell-1}$};
\node[func, align=center, minimum height=12mm] (causal_pad) [below=10mm of rational_func.center] {\small \tt pad};
\node[func, align=center, minimum height=12mm] (rfft_2) [below=10mm of rfft_1.center] {\small \tt rFFT};
\node[align=center] (elem_prod) [left=2.5mm of rfft_2] {$\odot$};
\node[func, align=center, minimum height=6mm] (irfft_2) [left=2.5mm of elem_prod] {\small \tt irFFT};
\node[params, text width=25mm, align=center] (ys) [below=10mm of irfft_2.center] {\small $y_0 \; \cdots y_{\ell-1} \; 0 \cdots 0$};

\draw [latex-] (rfft_1.west|-bs.east) -- (bs.east);
\draw [latex-] (rfft_1.west|-as.east) -- (as.east);

\draw [latex-] ([yshift=3.25mm]rational_func.west) -- ([yshift=3.25mm]rfft_1.east);
\draw [latex-] ([yshift=-3mm]rational_func.west) -- ([yshift=-3mm]rfft_1.east);
\draw [latex-] (irfft_1.west) -- (rational_func.east);
\draw [latex-] (hs.north) -- (irfft_1.south);
\draw [latex-] (causal_pad.east|-hs.west) -- (hs.west);
\draw [latex-] (causal_pad.east|-us.west) -- (us.west);
\draw [latex-] ([yshift=3.5mm]rfft_2.east) -- ([yshift=3.5mm]causal_pad.west);
\draw [latex-] ([yshift=-3mm]rfft_2.east) -- ([yshift=-3mm]causal_pad.west);
\draw [latex-] ([yshift=-1mm,xshift=-1mm]elem_prod.east) -- ([yshift=-3mm]rfft_2.west);
\draw [latex-] ([yshift=1mm,xshift=-1mm]elem_prod.east) -- ([yshift=3.5mm]rfft_2.west);
\draw [latex-] (irfft_2.east) -- ([xshift=1mm]elem_prod.west);
\draw [latex-] (ys.north) -- (irfft_2.south);



\node[text width=.48\textwidth, align=center, anchor=east] (seq_inf_heading) [right=0mm of par_inf_heading.east] {\textit{\textbf{Recurrent Form (c)}}};

\node[] (ssm) [below=5mm of seq_inf_heading] {\small
$\left[ 
    \begin{array}{c}
    \begin{matrix}
        x_{t+1}^{1} \\
        x_{t+1}^{2} \\
        x_{t+1}^{3} \\
        \vdots \\
        x_{t+1}^{n} \end{matrix} \\
        \hline
        \begin{matrix}
        y_t
    \end{matrix}
    \end{array}
\right] = \left[
        \begin{array}{c|c}
            \begin{matrix}
                -{\color{a}a_1} & -{\color{a}a_2}   &\cdots      &- {\color{a}a_n} \\
                1               & 0                 & \cdots     & 0 \\
                0               & 1                 & \cdots     & 0 \\
                \vdots          & \vdots            & \ddots     & \vdots\\
                0               & 0                 & \cdots     & 0
            \end{matrix} & \begin{matrix}1 \\ 0 \\ 0 \\ \vdots \\ 0\end{matrix}\\
            \hline
            \begin{matrix}~~{\color{b}b_1} & ~~~~{\color{b}b_2} & ~\cdots~ & ~{\color{b}b_n}\end{matrix} & {\color{h}h_0}
        \end{array} 
    \right]\left[
    \begin{array}{c}
    \begin{matrix}
        x_t^{1} \\
        x_t^{2} \\
        x_t^{3} \\
        \vdots \\
        x_t^{n} \end{matrix} \\
        \hline
        \begin{matrix}
        u_t
    \end{matrix}
    \end{array}
    \right]$};
\end{tikzpicture}
\vspace{-10mm}
\caption{\small
\textit{\textbf{(a)}} The rational transfer function (RTF) representation comprises numerator and denominator polynomial coefficients {\color{b}$\textbf{b}$} and {\color{a}$\textbf{a}$}, and the feedforward term {\color{h}$h_0$}. \textit{\textbf{(b)}} illustrates the proposed state-free parallel inference algorithm. The key to efficient state-free inference lies in casting {\color{b}$\textbf{b}$} and {\color{a}$\textbf{a}$} onto the sequence length for computing the convolutional filter $(h_i)_{i \in [\ell]}$. \textit{\textbf{(c)}} illustrates the recurrent form of RTF which can be used for fast single-step inference. Here we denote the $i$-th state at time $t$ as $x_t^{i}$.}  \label{fig:summary}
\end{figure*}
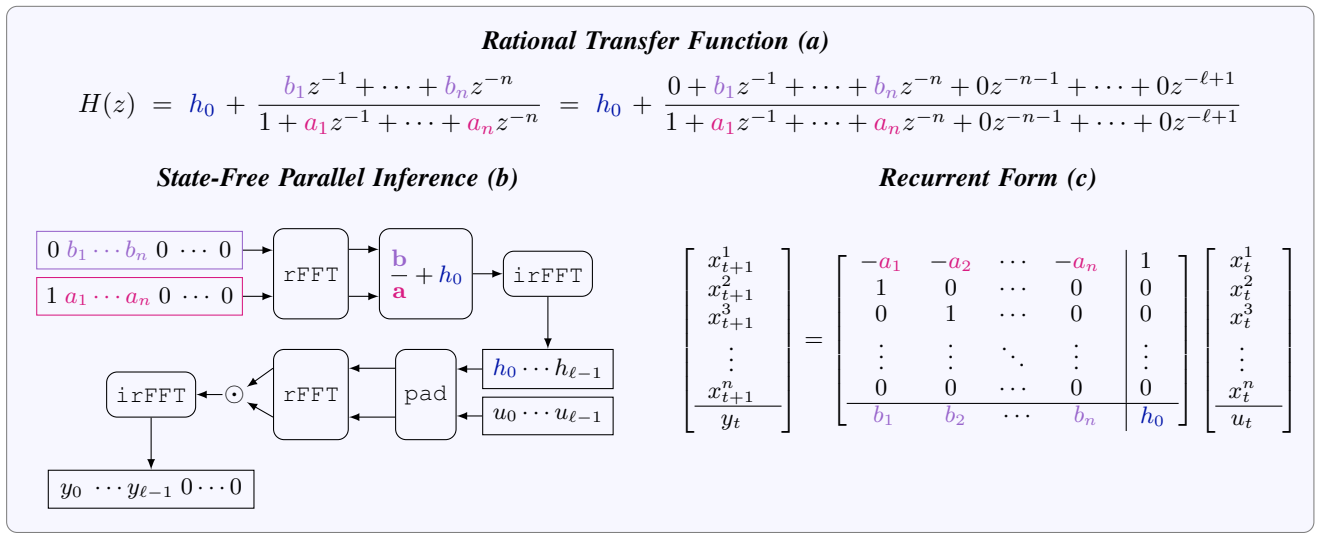

With the goal of enabling parallel inference, many algorithms such as S5 \cite{s5}, LRU \cite{lru} S4 \cite{s4} and DSS \cite{dss} employ a modal (diagonal) SSM representation, wherein the state transition matrix $\sA$ is diagonal, potentially limiting the model's expressive capacity for a given state dimension. Additionally, along with Mamba \cite{mamba}, S5 and LRU rely on the parallel scan directive \cite{parallelrnn, sum_prefix} which incurs considerable memory costs at large state sizes\footnote{Even when the states are only materialized in SRAM \cite{mamba}, as SRAMs are limited in size.}, due to the materialization of states over the sequence length, as made evident in Figure \ref{fig:mem}.

The expensive space requirement is alleviated with S4 \cite{s4}, S4D \cite{s4d}, and SpaceTime \cite{spacetime} by an algorithm that admits what we denote as \textbf{state-additive} space complexities, in which the parallel inference algorithm collapses the state dimension $n$ onto the sequence length dimension $\ell$, enabling space complexities of $\cO(\ell+n)$ in place of the much greater \textbf{state-multiplicative} $\cO(\ell n)$ complexity of scan-based algorithms. To realize the aforementioned state-additive space complexity, S4 and S4D leverage fast Cauchy and Vandermonde matrix-vector product algorithms \cite{structured_matrices_and_polynomials}. These algorithms used in computing the convolutional kernel for S4 and S4D scale as $\cO((\ell+n) \log ^2 (\ell+n))$, bottlenecking the faster $\cO(\ell \log \ell)$ required to execute the downstream convolution.

We approach solving these issues through a thorough frequency analysis of state-space models and unveil a parallel inference algorithm that admits \textbf{state-free} space and time complexities of $\cO(\ell)$ and $\cO(\ell \log \ell)$ respectively. Additionally, the proposed algorithm operates over a complete representation, the \textbf{Rational Transfer Function} (RTF) representation, which unlike diagonal SSMs \cite{s4d, dss, s5}, fully encapsulates the functional space of any linear time-invariant state-space model, including ones parameterized with dense matrices. Parallel inference with RTF solely relies on the Fast Fourier Transform ($\sf FFT$) algorithm -- a widely used and optimized algorithm, alleviating the need for additional custom low-level optimizations to obtain efficient subquadratic complexities. Figure \ref{fig:summary} illustrates an overview of the parametrization, parallel inference, and sequential inference algorithms of our proposed SSM.

In order to validate the proposed parametrization, we conducted experiments across a range of tasks, models, and importantly state sizes, including Long Range Arena (LRA), language modeling, and synthetic tasks. Notably, in LRA our proposed model obtained state-of-the-art accuracy (Table \ref{table:lra_accuracy}) among other attention-free models, and faster training speeds in comparison to S4 and S4D across state sizes (Figure \ref{fig:lra_speed}). We approached language modeling by embedding RTF into a Hyena model \cite{hyena}, effectively replacing the original convolutional filter parameterized with MLPs with transfer functions, and observed improved perplexity over the Hyena Filter baseline when trained on WikiText103 (Table \ref{table:wt103}).
\section{Preliminaries and Related Work}
We discuss sequence modeling, convolution-based sequence processing units and their state-space realization.
\subsection{Sequence Modeling with Convolutions}
Let $\bS_{\ell}^d$ denote the space of length-$\ell$ vector-valued sequences, $\bS_\ell \coloneqq \{(u_t)_{t \in [\ell]} : u_t \in \R^d \}\equiv \R^{\ell \times d}$. We denote the time index with a subscript roman letter and additional dimensions with greek superscripts, e.g. $x_t^\alpha$ for $t\in[\ell]$ and $\alpha \in [d]$. Any map from $\bS_\ell^d$ into itself is herein referred to as a \textit{sequence processor}. Complex \textit{deep} learning architectures tailored for sequence modeling typically involve the composition of simpler, parametric sequence processors in a multi-layer fashion. In this work, we focus on causal sequence processors $u \mapsto y$, where the output $y_t$ at any given time $t\in [\ell]$ is a function of solely the preceding inputs, i.e. $\partial y_t /\partial u_j = 0$ for all $t<j$ and $u \in \bS_\ell^d$. This constraint is crucial, for instance, in auto-regressive training of decoder-only language models \cite{unsupervisedllm} or analogous modeling tasks of temporal dynamics (see e.g. \citealp{decisiontransformer}). 

The ideal sequence processing layer is expected to fulfill several design criteria, balancing factors such as expressivity, computational and memory efficiency, favorable training dynamics, and parametric efficiency. Of particular interest in this work are those sequence processors that utilize \textit{single-input single-output} (SISO) discrete convolutions as their fundamental components, a.k.a. \textit{linear time invariant (LTI) systems}, with convolutional filters being implicitly parameterized. 
\begin{tcolorbox}[enhanced, frame hidden, sharp corners, colback=ghostwhite]
    A \textbf{single-input single-output causal convolution} between an \textit{input} $u \in \bS_\ell^1$ and a \textit{filter} $h \in \bS_\ell^1$ (often called the \textit{impulse response} function) is defined as
    \begin{equation}\label{eq:cnn}
        (h * u)_t = \sum_{j=0}^{t} h_{t -j} u_j\quad \text{for all }  t \in [\ell].
    \end{equation}
    The class of \textit{implicit} convolutions represent the filter as a parametric function $f_\theta: t \mapsto h_t \coloneqq f_\theta(t)$.
\end{tcolorbox}
SISO convolution operators can be represented by structured (Toeplitz) matrices that admit a fast multiplication algorithm with efficient sub-quadratic complexity $\cO(\ell \log \ell)$. They serve as the fundamental building blocks on various classical signal processing pipelines such as audio systems \cite{oppenheim99} and visual systems \cite{gonzalez2008digital}.

A notable modern example of sequence processors that make use of implicit convolutions as their core operation on the temporal dimension is the Hyena architecture \cite{hyena}. Given three sequences $q,k,v \in \bS_\ell^d$ obtained from the input $u\in\bS^d_\ell$ through three dense linear projections $\R^d\to\R^d$ followed by three \textit{short} convolutions, Hyena realizes a map $u \mapsto \cH u : \bS_\ell^d \to \bS_\ell^d$, defined element-wise for all $t \in [\ell]$ and $\alpha \in [d]$ as
\begin{equation} \label{eq:multisiso_hyena}
    \begin{aligned}
        {(\cH u)}_t^\alpha &= u_t^\alpha + \sum_{\beta=0}^{d-1}\sum_{j=0}^t \sT^{\alpha\beta}q_t^\beta {\color{h}h^\beta_{t-j}} k_j^\beta v_j^\beta
    \end{aligned}
\end{equation}
where $\{h^\alpha_t : t \in [\ell], \alpha \in [d]\}\in \bS^d_\ell$ is a collection of implicit \textit{long} convolution filters and $\sT\in\R^{d\times d}$ is an \textit{output} projection that mixes channels across the sequence length. Hyena applies $d$ SISO convolutions, independently on each channel. This \textit{multi} SISO approach has been successful in other convolution-based sequence processors such as S4 \cite{s4,s4d} or H3 \cite{h3}  (as well as linear input-varying models \cite{mamba}).
\subsection{State-Space Realization of Convolutions}
This work delves deep into the design of the individual SISO filters $h_t$, tailored for sequence processing architectures leveraging classical frequency-domain analysis techniques from signal processing and control theory. 

More specifically, we specialize on those filters that admit a finite-dimensional state-space (lumped) realization, i.e. the input-output relation of their induced convolution operator can be expressed as: 
\begin{equation}\label{eq:dtssm}
    \begin{aligned}
        x_{t+1} &= \sA x_t + \sB u_t\\
        y_t &= \sC x_t + h_0 u_t
    \end{aligned}~,\quad 
    t\mapsto
    h_t = 
    \begin{cases}
        h_0 & t=0\\
        \sC \sA^{t-1} \sB & t>0
	\end{cases}
\end{equation}
with a finite-dimensional \textit{state} $x_t \in \R^n$ ($n\ll \ell$), \textit{input} $u_t \in \mathbb{R}$, and \textit{output} $y_t \in \mathbb{R}$. Our trainable degrees of freedom are the matrices $\sA\in\R^{n \x n}$, $\sB\in\R^{n\x 1}$, $\sC\in\R^{1\x n}$, and $h_0\in\R$. The initial condition $x_0\in\R^n$ is usually set to zero such that $u\mapsto y$ is a pure convolution.
A major advantage of having a state-space realization is the possibility to switch between its convolution mode, for training, and recurrent mode, for efficient auto-regressive generation (see \citealp{lhyena} and Section \ref{sec:linsys} for further details and denominations). 
\paragraph{State-space representations} Parametrization of lumped convolutional filters with temporal dynamics, i.e., state-space parametrization present several challenges. Firstly, recurrence with dense transition matrices $\sA$ are computationally expensive, amounting to a computational complexity of $\cO(\ell n^2)$. To make such systems feasible various recent works proposing efficient state-space models have resorted to diagonalization \cite{s4d, s5, lru} and low-rank add-ons \cite{s4} of $\sA$.
As will be further uncovered when analyzing the dual representation, transfer functions, these restrictions impose a constraint on the expressivity of its convolutional filter $h$, given a fixed state-size $n$. 
Moreover, despite various works on optimizing parallel inference efficiency, associative scans utilized in \cite{parallelrnn, s5, lru, mamba} still incur considerable memory costs due to its state-multiplicative complexity of $\cO(\ell n)$, whereas fast Cauchy and Vandermonde matrix-vector products \cite{structured_matrices_and_polynomials} utilized in \cite{s4, s4d} present an improved state-additive space complexity of $\cO(\ell + n)$, but heavily rely on custom platform specific low-level optimizations. %
\section{Training SSMs in the frequency domain}

Linear time-invariant dynamical systems \eqref{eq:cnn} are completely characterized by their \textbf{impulse response} $h$, and in the case they admit a state-space realization \eqref{eq:dtssm}, their system matrices $(\sA,\sB,\sC,h_0)$.
\subsection{Transfer Function Representation}
An alternative \textbf{complete representation} of \eqref{eq:dtssm} is its \textit{transfer function} $H:\bC \to \bC$, defined as the $\cZ$-transform of the impulse response $H(z) \coloneqq \sum_{t\in\bN}h_t z^{-t}$ for all $z {\in} \bC$ where the sum converges. The transfer function of a state-space model $(\sA,\sB,\sC,h_0)$ is a \textit{proper\footnote{i.e. such that the denominator's order is not less than the numerator's one.} rational function} of $z$,
\begin{equation}\label{eq:dtssm_tf}
    \begin{aligned}
        H(z) &= h_0 + \sC(z\sI-\sA)^{-1}\sB \\
        &= h_0 + \frac{b_1z^{-1} + ~\cdots~ + b_n z^{-n}}{1 + a_1z^{-1} + ~\cdots~ + a_nz^{-n}}.
    \end{aligned}
\end{equation}
Refer to \ref{sec:tf_realization} for complete derivations.
As discrete convolutions are the dual operation to element-wise multiplication under $\cZ$-transform, the input-output relation of any LTI system can be equivalently characterized by $H(z)$,
\[
    y_t = (h * u)_t ~~\Leftrightarrow~~ Y(z) = H(z)U(z)
\]
where $H$ is defined outside the circle in the complex plane whose radius is the amplitude of the largest eigenvalue of the state transition matrix $\sA$.  The $\cZ$-transform is a projection of the sequence onto a power basis $z^{-t} = r^{-t}e^{-i\omega t}$ for $r,\omega\in \R$. This basis is not orthogonal unless $r=1$. That is the basis of the discrete-time Fourier transform $\cF$. Hence, the discrete-time Fourier transform of the signal $h$ is defined as $\cF[h](e^{i\w})=H(e^{i\w})\coloneqq\sum_{t\in\bN}h_t e^{-i\w t}$, i.e. it is the transfer function $H(z)$ evaluated at $z=e^{i\omega}$. We say that sequences live in the \textit{time domain} and their $\cZ$ (or $\cF$) transforms in the \textit{frequency domain}.

We argue that parametrizing state-space models via their transfer function (i.e. making $(a,b)$ the \textit{learnable} parameters), encompasses previous representations of SSMs such as using structured matrices \cite{h3,s4} or \textit{modal canonical forms} \cite{s4d,lru,s5,h3}.

\paragraph{Coordinate invariance of the transfer function} 
Notably, the transfer function is an \textit{invariant} of the system: if an invertible change of variables is applied to the state-space representation, the transfer function parameters $(a,b)$ remain unchanged. Without loss of generality let $h_0=0$.
\begin{tcolorbox}[enhanced, frame hidden, sharp corners, boxsep=0pt, before skip=0pt, after skip=0pt, colback=ghostwhite]
\begin{lemma}\label{prop:tf_invariant}
    Coefficients $a,b$ are \textbf{invariant} under any invertible change of variables.
\end{lemma}
\end{tcolorbox}
\proof The proof is classic and can be found in \cite{chenlinearsystem} and follows from the definition of \textit{equivalence transformation}. Consider the state-space matrices under a change of variables $\hat x = \sK x$, for some invertible $\sK \in\R^{n \times n}$
$$
\hat \sA = \sK\sA\sK^{-1},~~\hat \sB = \sK\sB,~~ \hat \sC = \sC\sK^{-1}.
$$
The transformed transfer function $\hat H(z)$ is given by 
$$
    \hat H(z) = \sC\sK^{-1}[\sK(z\sI - \sA)\sK^{-1}]^{-1}\sK\sB = H(z)
$$
\endproof
This emergent coordinate invariance should be of warning to most attempts at modeling filters by directly learning either {dense} or structured state-space matrices $(\sA,\sB,\sC)$ as such: there are infinitely many equivalent state-space realizations that map to the same system. This also demonstrates that dense SSM parametrizations are inefficient in their use of parameters with respect to its expressivity.

\paragraph{Expressivity of the transfer function} Any impulse response $h$ that can be represented using dense matrices—of $n^2+2n+1$ parameters with stable dynamics—can also be described using rational transfer functions with just $2n+1$ parameters.

This is demonstrated in the derivations presented in Section \ref{sec:ssm2tf}. It illustrates that one can calculate the parameters of the transfer functions $(a,b,h_0)$, given any state-space parameterization $(\sA, \sB, \sC, h_0)$, through the following method:
\begin{equation}\label{eq:ssm2tf_main}
\begin{aligned}
a &= {\sf poly}({\sf eig}(\sA)), \\
b &= {\sf poly}({\sf eig}(\sA - \sB\sC)) + {\sf poly}({\sf eig}(\sA))(h_0 - 1),
\end{aligned}
\end{equation}
in which ${\sf poly}(r)$ computes the coefficients of a polynomial given its roots $r_0, \dots, r_n$.

Parallel to change of variable techniques such as diagonalization of $\sA$ employed in time-domain state-space realizations, partial fraction decomposition of transfer functions can not only provide alternative representations of state-space models, but also intuitive insights on the expressivity of these models. 

As an example, by simply taking the first order partial fraction decomposition of a rational transfer function $H(z)$, i.e.,
\begin{equation}\label{eq:tf_modal_main}
    H(z) = \sum_{i = 1}^{n}{\frac{r_i}{z-\lambda_i}} + h_0
\end{equation}
in which $r_i,\lambda_i \in \bC$, we obtain the diagonal time-domain parameterization. Its equivalence can be shown by simply breaking down the geometric series $r_i/(z-\lambda_i) = r_i(1/z + \lambda_i/z^2 + \lambda_i^2/z^3 + \dots)$, and applying the inverse $\mathcal{Z}$-transform ($z^{-j}$ is an impulse at time-step $j$), resulting in the diagonal SSM convolutional kernel $h_t = \sum_{i\in[n]}r_i \lambda_i^{t-1}$ for $t >0$.

Looking further, we observe that, like (\ref{eq:dtssm_tf}), it contains $2n+1$ trainable parameters, but does not permit repeated roots, i.e. $r_1/(z-\lambda_1) + r_2/(z-\lambda_1)^2$, thereby demonstrating its limited expressivity.

\subsection{State-Free Parallel Inference}

\begin{algorithm}[tb]
   \caption{RTF Kernel Generation}
   \label{alg:kernel}
\begin{algorithmic}
   \STATE {\bfseries Input:} RTF params $(a,b,h_0)$, truncation length $\ell$ 
   \STATE $\bar{b},\; \bar{a} \gets \texttt{pad}(b, a, (1,\ell - n-1))$ \hfill {\small \color{mayablue}\# Padding $a$ and $b$ to $\ell$}
   \STATE $\bar{a}_0 \gets 1$  \hfill {\small \color{mayablue} \# Set denominator monic poly. term.}
   \STATE $B,\; A \gets \texttt{FFT}_{\ell}(\bar{b}, \bar{a})$ \hfill {\small \color{mayablue} \# Polynomial eval.}
   \STATE $H \gets B/A + h_0$ \hfill {\small \color{mayablue} \# Construct rational function}
\end{algorithmic}
\end{algorithm}

For attaining sub-quadratic parallel inference speeds, the approach taken by S4, S4D, and SpaceTime predominantly hinges on the efficient computation of its length-$\ell$ truncated impulse response $h_t$:
\begin{equation}
    h_t = \left\{
    \begin{matrix*}[l]
        h_0 & t=0 \\
        \sC \sA^{t-1} \sB & 0< t\leq\ell\\
        0 & t>\ell
    \end{matrix*}
    \right.
    ,
\end{equation}
or its corresponding spectrum $\mathsf{FFT}_\ell(h)$ for downstream integration with the sub-quadratic convolution algorithm, ${\sf FFTConv}(u, h)$, described in \cite{burrus1985convolution, selesnick2017fast, flashfftconv}.

Adopting a parallel approach for rational transfer function, we reveal that $h_t$ can be computed in a state-free manner, incurring space and time complexities of $\cO(\ell)$ and $\cO(\ell \log \ell)$, respectively. This is achieved through the evaluation of the \textbf{truncated transfer function} $H_\ell(z)$ across the roots of unity, as delineated below.

Firstly, we demonstrate that an impulse response of length-$\ell$, when expressed in the $\cZ$-domain as $H_\ell(z) = \sum^{\ell -1}_{t=0}h_tz^{-t}$, can be efficiently transformed into its time-domain representation in the following manner.
\begin{tcolorbox}[enhanced, frame hidden, sharp corners, boxsep=0pt, before skip=0pt, after skip=0pt, colback=ghostwhite]
\begin{lemma}\label{lemma:inv_z}
    Let $\bT_m$ denote the set of the $m$ roots of unity, i.e. $\bT_m \coloneqq \{z^k : z = e^{2\pi i/m}\}_{k \in[m]}$. Then, for all $t \in [\ell]$ and $m\geq \ell$ it holds
    \begin{equation}
        h_t = \mathsf{iFFT}_m\big((H_\ell(z))_{z \in \bT_m}\big)_t \label{eq:get_ht}
    \end{equation}
\end{lemma}
\end{tcolorbox}
\begin{proof}
\begin{equation}
\begin{aligned}
    \mathsf{iFFT}_m\big((H_\ell(z))_{z \in \bT_m}\big)_t &= \frac{1}{m}\sum_{z \in \bT_m} H_\ell(z) z^t \\  
    &= \frac{1}{m}\sum_{z \in \bT_m} \sum_{j=0}^{\ell-1}{h_j z^{t-j}} \label{eq:z_inv} \\
    &= \frac{1}{m}\sum_{j=0}^{\ell-1}h_j \begin{cases}
    m & t-j=0\\
    0 & \text{otherwise}
    \end{cases}\\
    &= h_t.
\end{aligned}
\end{equation}
\end{proof}
Additionally, observe that the inverse application of Lemma \ref{lemma:inv_z} results in the following insight.
\begin{tcolorbox}[enhanced, frame hidden, sharp corners, colback=ghostwhite]
\textbf{Evaluating a truncated transfer function} $H_\ell(z)$ at the roots of unity, outputs the spectrum of the impulse response, that is: \begin{equation} (H_\ell(z))_{z \in \bT_m} = \mathsf{FFT}_m(h).\end{equation}
\end{tcolorbox}
In order to truncate the rational transfer function, we devise a ``tail'' $\tilde{H}_\ell(z)$, such that $H_\ell(z) = H(z) - \tilde{H}_\ell(z)$, as follows.
\begin{tcolorbox}[enhanced, frame hidden, sharp corners, boxsep=0pt, before skip=0pt, after skip=0pt, colback=ghostwhite]
\begin{lemma}\label{lemma:trunc_tf}
Let the ``tail'', $\tilde{H}_\ell(z)$ be a $\cZ$-domain representation a lumped LTI system $(\sA, \sB, \sC, h_0)$ for $t>\ell$, i.e. $\tilde{H}_\ell(z) = \sum_{t = \ell+1}^{\infty}\sC\sA^{t-1}\sB z^{-t}$, then
\begin{equation}
\tilde{H}_{\ell}(z) = \sC \sA^{\ell}z^{-\ell}(z\sI - \sA)^{-1} \sB.
\end{equation}
\end{lemma}
\end{tcolorbox}
\begin{proof}
\begin{equation}
\begin{aligned}
\sum_{t = \ell+1}^{\infty}\sC\sA^{t-1}\sB z^{-t} &= \sC \sA^{-1} \left[ \sum_{t=\ell+1}^{\infty}\sA^t z^{-t} \right]\sB \\
&= \sC \sA^{-1} \left[ \sA^{\ell+1}z^{-\ell-1}(\sI - \sA z^{-1})^{-1} \right]\sB \\
&= \sC \sA^{\ell}z^{-\ell-1}(\sI - \sA z^{-1})^{-1} \sB \\
&= \sC \sA^{\ell}z^{-\ell}(z\sI - \sA)^{-1} \sB.
\end{aligned}
\end{equation}
\end{proof}

Since $z^{-\ell} = 1 \forall z \in \bT_\ell$, we can derive the length-$\ell$ truncated transfer function in the following manner,
\begin{equation}
\begin{gathered}
    H_\ell(z) = H(z) - \tilde{H}_\ell(z) = \tilde{\sC}(z\sI - \sA)^{-1}\sB, \\
    \tilde{\sC} = \sC(\sI - \sA^\ell) \coloneqq \tilde{b},
\end{gathered}
\end{equation}
Nonetheless, in practice, we circumvent the computation of $\sA^{\ell}$, by directly optimizing $\tilde{b}$ during the training phase, and only apply the inverse correction $\sC = \tilde{b}(\sI - \sA^{\ell})^{-1}$, upon deployment, i.e. autoregressive inference. This is equivalent to the approach taken by \cite{s4, s4d, spacetime}, on the ``truncated SSM generating function''.

To evaluate the truncated rational function, we recognize that:
\begin{enumerate}[topsep=0pt]
    \item Rational functions are composed of polynomials. 
    \item Evaluating polynomials on the roots of unity, is equivalent to applying a fast Fourier transform over its coefficients.
\end{enumerate}
\begin{tcolorbox}[enhanced, frame hidden, sharp corners, boxsep=0pt, before skip=0pt, after skip=0pt, colback=ghostwhite]
\begin{lemma}\label{lemma:poly_eval}
Let $\alpha_k$ be the $k$-th order coefficient of a polynomial. Then for all $k,t \in [m]$, $z = e^{2\pi i /m}$, it holds
\begin{equation}
    \sum_{k=0}^{m-1} \alpha_k z^{-tk} = {\sf FFT}_m (\alpha)_t,
\end{equation}
\end{lemma}
\end{tcolorbox}
\begin{proof}
    By definition of the Fourier Transform.
\end{proof}
In light of Lemma \ref{lemma:poly_eval}, it becomes evident that for any $n$-th order truncated rational transfer function parameterized by $(a, \tilde{b}, h_0)$, by setting $a_0 = 1$, $\tilde{b}_0 = 0$ and $a_k, \tilde{b}_k = 0$ for $k>n$ (zero padding of polynomial coefficients), the spectrum of the impulse response can be computed with:
\begin{tcolorbox}[enhanced, frame hidden, sharp corners, boxsep=0pt, before skip=0pt, after skip=0pt, colback=ghostwhite,]
\begin{equation}
    H_\ell(z^t) = \frac{\sum_{k=0}^{\ell-1}\tilde{b}_k z^{-tk}}{\sum_{k=0}^{\ell-1}a_k z^{-tk}} + h_0 =\frac{\mathsf{FFT}_\ell(\tilde{b})_t}{\mathsf{FFT}_\ell(a)_t} + h_0,
\end{equation}
\end{tcolorbox}
as demonstrated in Algorithm \ref{alg:kernel}.
Finally to obtain $h_t$, we simply apply Equation \ref{eq:get_ht}.

Importantly, the proposed parallel inference algorithm relies solely on the $\mathsf{FFT}$ algorithms, which have space and time complexities of $\mathcal{O}(\ell)$ and $\mathcal{O}(\ell \log \ell)$, respectively. The ubiquitous $\mathsf{FFT}$ algorithm is widely used and already have low-level optimizations applied across several platforms, subsequently optimizing RTF across those platforms.

\subsection{Fast Companion Recurrence}

Rational transfer functions could directly be translated into a structured state-space model of the following form:
\begin{equation} \label{eq:companion_ssm}
\begin{gathered}
    x_{t+1} =
    \begin{bmatrix}
        -a_1 & -a_2 & \cdots & -a_n \\
        1 & 0 & \cdots & 0 \\
        0 & 1 & \cdots & 0 \\
        \vdots & \vdots & \ddots & \vdots \\
        0 & 0 & \cdots & 0
    \end{bmatrix}
    x_t + 
    \begin{bmatrix}
        1 \\
        0 \\
        0\\
        \vdots \\
        0
    \end{bmatrix} u_t \\
    y_t = \begin{bmatrix}
        b_1 & b_2 & \cdots & b_n 
    \end{bmatrix} x_t + h_0 u_t.
\end{gathered}
\end{equation}

The structure (companion form) permits fast companion recurrence via the combination of shift operations, dot products resulting in single time-step space and time complexities of $\cO(n)$. Refer to Section \ref{app:companion_recurrence} for the full derivation.

Moreover, as discussed in \cite{lhyena}, the companion realization of a state-space model can be leveraged to perform fast \textit{prefilling}, in which the state $x_t$ can be obtained from $u_0, \dots, u_t$ with computation complexity of $\cO(\ell \log_2 \ell)$. Fast prefilling is applicable in extensive language modeling applications, where the model, upon receiving a length-$\ell$ prompt from the user, autoregressively generates subsequent prompts using a constant-time recurrent algorithm as described above. For state-space realizations that are not in companion form, they must first be transformed into the companion form using Equation \eqref{eq:ssm2tf_main} to perform fast prefilling.

Unlike SpaceTime \cite{spacetime} that shares the same $\sA$ matrix but trains both $\sB$ and $\sC$, we adhere to the true companion form during training, in which the $\sB$ matrix is a constant as shown in Equation (\ref{eq:companion_ssm}), while $b$ ($\sC$ matrix) is trained.


\subsection{Stable Parametrization}

To prevent numerical instabilities, it is important to configure SSMs to exhibit stable dynamics. The choice of parameters for the state-transition matrix $\sA$ significantly influences their stability. For rational transfer functions, the roots of the denominator polynomial (the pole) must lie within the complex unit circle, i.e. $|r| \leq 1$ to prevent unstable dynamics \cite{chenlinearsystem}.

Unlike diagonal SSMs, with first order roots (Equation (\ref{eq:tf_modal_main})), ensuring that the coefficients of a high order polynomial $\sum_{i=0}^{n}{a_{n-i} z^{i}}$ are such that its roots remain within the complex unit circle presents a complex challenge, as highlighted in \cite{constrain_roots}. SpaceTime \cite{spacetime} adopts Montel's method \cite{montel, constrain_roots}, a technique that, for a Monic polynomial (where $a_0 = 1$), constrains the remaining coefficients in a manner described by:
\begin{equation}
\sum_{i=1}^{n-1}{|a_i|} \leq 1. \label{eq:Montel}
\end{equation}

However, as depicted in Figure \ref{fig:roots}, the application of Montel's method not only ensures that the roots are confined within the unit circle but also limits them to a specific subset of the stable region. This limitation could potentially diminish performance, a phenomenon supported by the findings in Table \ref{table:wt103-constrain}.

To mitigate this, we propose an alternative initialization strategy for the SSM coefficients, aiming to position them as far as possible from violating Montel's constraint:
\begin{equation}
\argmin_{a}(\sum_{i=1}^{n-1}{|a_i|}) = \mathbf{0},
\end{equation}
where $a,\tilde{b} = \mathbf{0}$. We denote this initialization scheme as the \textbf{zero} initialization. Our ablation tests (Table \ref{table:wt103-constrain}) and comparisons against SpaceTime on the Long Range Arena \cite{lra} benchmark (Table \ref{table:lra_accuracy}) show enhanced training stability and consequently, improved performance when adopting the zero initialization scheme.\footnote{Unless explicitly stated otherwise, all results presented in this paper adopts the zero initialization scheme with $h_0 = 1$.}

\section{Experimental Results}

In this section, we conduct an empirical evaluation of RTF in comparison to other state-space models and sequence models. Section \ref{exp:mem} is dedicated to assessing memory usage and processing speed. Sections \ref{exp:lra} and \ref{exp:synthetics} examine the ability for SSMs to memorize and model long-range dependencies. Finally, their ability to model language is assessed in sections \ref{exp:lh} and \ref{exp:wt103}.

\subsection{Efficiency Profiling} \label{exp:mem}

We profiled GPU memory usage between a parallel scan-based S5 model and RTF across different sequence lengths and state sizes at channel dimensions of $d=1024$. The results depicted in Figure \ref{fig:mem} reveal a consistent trend, wherein the memory consumption for the scan operation rises in conjunction with state size and sequence length, while it solely escalates with sequence length for RTF. This phenomenon can be attributed to the aforementioned state-free characteristic of RTF's inference algorithm, which casts its parameters with size of the state dimension onto the sequence length for parallel inference. We also observed a similar trend for the inference latency which is further detailed in Appendix \ref{app:memory}. 

\begin{figure}
    \centering
    \includegraphics[trim={0 3mm 0 0},clip, width=.45\textwidth]{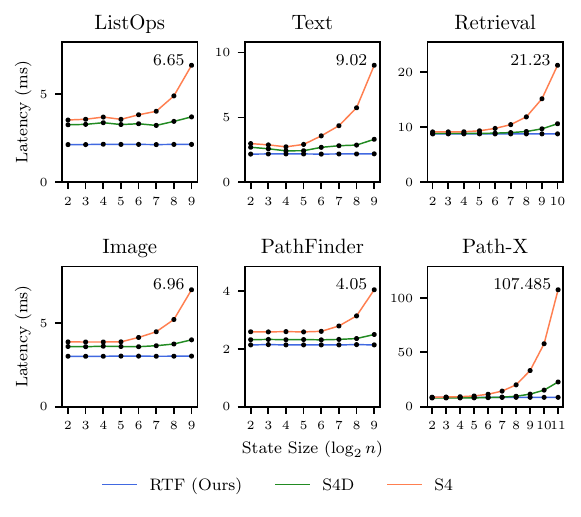}
    \caption{Latency profiles for a single RTF, S4D, and S4 layer at various state sizes. It is evident that RTF consistently exhibits superior parallel inference speeds, with its lower latency across a range of tasks and state sizes.}
    \label{fig:lra_speed}
\end{figure}

Next, we profiled inference latency across different SSMs of varying state-sizes over a suite of six LRA tasks, facilitating speed comparisons across a wide range of model architectures. Figure \ref{fig:lra_speed} reports the median inference latency per SSM layer across 75 training iterations.

The results show a recurring trend, wherein RTF's inference latency remained consistent regardless of state size and conversely, S4D and S4 experienced slower speeds particularly at higher orders, due to the utilization of the slower Vandermonde or Cauchy matrix-vector product algorithms respectively, which have computational complexity of $\mathcal{O}((\ell + n) \log ^2 (\ell + n))$ as opposed to RTF's $\cO(\ell \log \ell)$.

\subsection{Modeling Long Range Dependencies} \label{exp:lra}

\begin{table*}
    \small
    \centering
    \setlength{\tabcolsep}{4pt}
    \caption{Long range arena benchmark results. We included results reported in \cite{s4, s5, ren2023sparse} and additionally ran SpaceTime \cite{spacetime} based on the official implementation with hyperparameters identical to RTF. We also included results of self-pretrained (SPT) Transformers \cite{transformer_lra} denoted with + Causal SPT. $^\dag$ indicate the use of an increased state-size and \xmark $\;$indicates that the model was unable to train beyond a random guessing policy.}
    \begin{tabular}{@{}l|ccccccc@{}}    
    \toprule
    Model & ListOps & Text & Retrieval & Image & Pathfinder & Path-X & Avg.\\
    \midrule 
    Transformer & $36.37$ & $64.27$ & $57.46$ & $42.44$ & $71.4$ & \xmark & 53.66 \\ 
    Luna-256  & $37.25$ & $64.57$ & $79.29$ & $47.38$ & $72.72$ & \xmark & 58.54 \\ 
    Transformers + Causal SPT & 59.15 & 88.81 & 90.38 & 76.0 & 88.49 & 88.05 & 81.81 \\
    Mega $\mathcal{O}(\ell^{2})$  & $\mathbf{63.14}$ & \textbf{90.43} & $91.25$ & \underline{90.44} & \underline{96.01} & \underline{97.98} & \textbf{88.21} \\ 
    H3 & 57.5 & 88.2 & 91.0 & 87.3 & 93 & 91.8 & 84.8 \\
    CCNN  & $43.6$ & $84.08$ & \xmark & $88.9$ & $91.51$ & \xmark & 68.02 \\ 
    Liquid-S4 & \underline{62.75} & 89.02 & 91.2 & 89.5 & 94.8 & 96.66 & 87.32\\
    \midrule 
    S5  & 62.15 & 89.31 & $91.4$ & $88.0$ & $95.33$ & $\mathbf{98.58}$ & 87.46 \\ 
    S4  & 61.29 & 88.25 & $90.90$ & 89.2 & $94.2$ & $96.35$ & 86.69 \\ 
    S4D & 60.74 & 87.03 & 90.68 & 89.18 & 95.42 & 97.32 & 86.72 \\ 
    SpaceTime & 56.4 & 87.8 & \underline{91.45} & 86.27 & \xmark & \xmark & 70.32 \\
    RTF (Ours) & 61.59 & \underline{89.72} & \textbf{92.04} & \textbf{90.51} & \textbf{96.11} & $96.32^{\dag}$ & \underline{87.71} \\ 
    \bottomrule 
    \end{tabular}
    \label{table:lra_accuracy}
\end{table*}

The Long Range Arena (LRA) benchmark has become a common ground for testing various sequence models including SSMs \cite{s4, s4d, s5, ls4} and Transformers \cite{transformer, performer}. It is composed of six classification tasks with long range input sequences of lengths ranging from $1024$ to $16384$. We conducted these experiments on RTF along with S4, S4D, and SpaceTime \cite{spacetime} as presented in Table \ref{table:lra_accuracy}.

RTF obtained strong results in several LRA tasks, including attaining state-of-the-art performance on \texttt{Retrieval}, and among attention-free approaches, the average score. However for Path-X, RTF was unable to learn a policy beyond random guessing when the state-size was fixed to 64, prompting an increase to 2048. Nevertheless, due to RTF's state-free parallel inference algorithm, this increase in state-size did not impact GPU memory consumption nor training speed as evidenced in Figure \ref{fig:lra_speed}.

\subsection{Synthetic Memorization Tasks} \label{exp:synthetics}

Recurrences have traditionally struggled with vanishing and exploding gradients, making memorization tasks challenging \cite{bengio_vanishing, vanishing}. To evaluate the memorization capabilities of our state-space model, we benchmark them against two synthetic memorization tasks: \textit{Copying} and \textit{Delay}.

The \textit{Copying} task, akin to \cite{urnn}, presents SSMs with 1024 length sequences of 64 discrete states sampled uniformly, which the model is then tasked to recall all 1024 tokens in order. Each model was given 10k training samples for 50 epochs, and was tested with 1000 unseen samples.

The \textit{Delay} task, which was also used to ablate HiPPO SSM initialization schemes \cite{hippo}, simply tests the model's ability to delay a continuous white noise by 1000 time steps. As reported by \citeauthor{hippo}, LSTMs and Transformers struggle on this seemingly simple task, and are unable to improve beyond a random guessing policy. The primary distinction between \textit{Copying} and \textit{Delay} is whether the input data is discrete or continuous. More detailed experimental setup could be found in \ref{app:synthetic}.

From the results reported in Table \ref{table:synthetics}, we observed that at higher state-sizes, RTF could more accurately copy and delay data. S4 on the other hand struggled on \textit{Copying}, showing no improvements beyond the state-size of 256. It is also worth noting that on both synthetic tasks, unlike the discrete-time RTF SSM, S4, being continuous-time required careful consideration of the initialization and interplay between the time-constant $\Delta$ and the transition matrix $\sA$ for reasonable performance. 

\begin{table}
    \small
    \centering
    \setlength{\tabcolsep}{4pt}
    \caption{Results on synthetic memorization tasks. The state-size of the model is denoted with the number trailing the model name, i.e. S4-64 is an S4 model with $n=64$.}
    \begin{tabular}{@{}l|cccc@{}}
    \toprule
    Model & \textit{Copying} & \textit{Delay} \\
    & acc. $\uparrow$ & RMSE $\downarrow$ \\
    \midrule 
    S4-64 & \textbf{29.3}     & \textbf{0.41} \\
    RTF-64 & 22.1             & 0.45 \\
    \midrule
    S4-128 & 34.2             & \textbf{0.39} \\
    RTF-128 & \textbf{93.3}   & 0.45 \\
    \midrule
    S4-256 & 35.0             & \textbf{0.33} \\
    RTF-256 & \textbf{100}      & 0.44 \\
    \midrule
    S4-512 & 33.1                  & $\mathbf{0.22}$ &\\
    RTF-512 & \textbf{100}                 & 0.38 \\
    \midrule
    S4-1024 & 33.2                 & 0.029 \\
    RTF-1024 & \textbf{100}                & $\mathbf{0.006}$  \\
    \bottomrule    
    \end{tabular}
    \label{table:synthetics}
\end{table}

\subsection{Laughing Hyena Distillation} \label{exp:lh}

\begin{table*}[t]
    \small
    \centering
    \setlength{\tabcolsep}{4pt}
    \caption{This table illustrates downstream evaluation scores from LM-Evaluation-Harness \cite{eval-harness}. The number trailing the model names indicate its state-size.}
    \begin{tabular}{@{}l|cccccccc@{}}
    \toprule
    Model & Winogrande & PIQA & HellaSwag  & OpenbookQA  & Distillation\\
    & acc. $\uparrow$ & acc. $\uparrow$ & acc. norm. $\uparrow$ & acc. norm. $\uparrow$ & MSE $\downarrow$\\ 
    \midrule 
    Baseline (160M) & 52.09 & 61.64 & 29.68 & 29.4  & - \\
    \midrule  
    LH-4 & 51.7 & 62.02 & 29.76 & 29.6  & 0.032\\ 
    RTF-4 & 51.7 & 61.04 & 29.82 & 29.6 & \textbf{0.018} \\ 
    \midrule  
    LH-16  & 52.25 & 61.75 & 29.73 & 28.6 & \textbf{0.009} \\ 
    RTF-16    & 52.96 & 61.64 & 29.85 & 29.8 & 0.013\\ 
    \midrule  
    LH-64  & 49.57 & 61.59 & 29.8 & 29.6  & \textbf{0.007} \\ 
    RTF-64    & 53.43 & 61.81 & 29.85 & 29.2 & 0.011 \\ 
    \bottomrule 
    \end{tabular}
    \label{table:distillation}
\end{table*}

Hyena \cite{hyena} and MultiHyena \cite{lhyena} operators utilize a diverse array of filters, encompassing short convolutional filters - filters implicitly parameterized by multi-layer perceptrons (MLP) \cite{hyena, siren, ckconv}, and diagonal SSMs \cite{lhyena}. Notably, Hyena operators with MLP-parameterized filters have demonstrated superior performance compared to other convolutional and recurrent methods, as highlighted in \cite{akyürek2024incontext, bhattamishra2023understanding}. Despite their effectiveness, these filters lack constant-time autoregressive inference speeds desired in applications such as language modeling. This limitation has led to the investigation of distilling MLP-based filters into SSMs, a process detailed in Laughing Hyena \cite{lhyena}.

Here, we look into distillation of MLP-based filters, using a 160M parameter multi-head StripedHyena \cite{poli2023stripedhyena} language model, trained on The Pile \cite{pile}, and compare distillation performances between RTF and a diagonal SSM employed in Laughing Hyena (LH), both of which boast highly efficient $\cO(n)$ autoregressive algorithms. Table \ref{table:distillation} reports distillation errors and downstream LM-Evaluation-Harness scores \cite{eval-harness}.

Interestingly, despite the theoretically superior expressiveness of RTF models, we observed that the modal representation employed in LH exhibits more favorable training dynamics for distillation at state-sizes 16 and 64, as evidenced by the distillation MSE. However with $n=4$, RTF outperforms LH while maintaining comparable downstream evaluation performances to the baseline model, making it a good candidate for unlocking efficient constant-speed autoregressive inference on Hyena language models.

\subsection{WikiText103 Language Modeling} \label{exp:wt103}

In addition to evaluating the language modeling capabilities of state space models through distillation techniques, their performance when directly trained on autoregressive cross-entropy loss \cite{unsupervisedllm} was investigated on the well-established WikiText-103 dataset. We used a Hyena operator and replaced its filters with RTF, which we refer to as \textit{Hyena-RTF}.

As shown in Table \ref{table:wt103}, Hyena-RTF outperforms both the Transformer and Hyena baselines on WikiText103. Additionally, RTF without the Hyena operator structure was compared against S4 and S4D on a pilot experiment further described in Appendix \ref{app:wt-pilot}, which similarly indicated relatively strong language modeling capability among other LTI SSMs. These results signal a promising potential for further scaling RTF on larger models and datasets.

\begin{table}
    \small
    \centering
    \setlength{\tabcolsep}{4pt}
    \caption{
    WikiText103 language modeling perplexity scores. The results are taken from \cite{hyena}. Each model listed below contains \texttt{$\sim$}125M parameters.
   }
    \begin{tabular}{@{}l|cc@{}}
    \toprule
    Model & Perplexity $\downarrow$ \\
    \midrule
    Transformer & 18.6\\
    Hybrid H3 & 18.5\\
    Linear Attention & 25.6\\
    Hyena & 18.5\\
    Hyena-S5 \cite{s5} & 18.3 \\
    Hyena-RTF (Ours) & \textbf{18.0} \\
    \bottomrule    
    \end{tabular}
    \label{table:wt103}
\end{table}

\section{Conclusion}

In this study, we explore \textit{state-space model} (SSM) parametrization via their dual representation, transfer functions. We systematically unveiled the realization of SSMs through rational transfer functions (RTF), demonstrating state-of-the-art efficiency through a state-free parallel inference algorithm, while maintaining the expressiveness of a dense SSM. Our experiments revealed that RTFs are effective for modeling long-range dependencies and processing language, and also exhibits improvements in comparison to the S4 model across synthetic memorization tasks with higher state-sizes. The results of our investigation suggest that RTFs hold significant potential for modeling signals across a variety of other domains.

\section{Acknowledgements}

T.S. was partially supported by JSPS KAKENHI (20H00576) and JST CREST (JPMJCR2015).  

\bibliography{references}

\begin{thebibliography}{59}
\providecommand{\natexlab}[1]{#1}
\providecommand{\url}[1]{\texttt{#1}}
\expandafter\ifx\csname urlstyle\endcsname\relax
  \providecommand{\doi}[1]{doi: #1}\else
  \providecommand{\doi}{doi: \begingroup \urlstyle{rm}\Url}\fi

\bibitem[Akyürek et~al.(2024)Akyürek, Wang, Kim, and Andreas]{akyürek2024incontext}
Akyürek, E., Wang, B., Kim, Y., and Andreas, J.
\newblock In-context language learning: Architectures and algorithms, 2024.

\bibitem[Alomari \& Chesneau(2022)Alomari and Chesneau]{constrain_roots}
Alomari, M.~W. and Chesneau, C.
\newblock Bounding the zeros of polynomials using the frobenius companion matrix partitioned by the cartesian decomposition.
\newblock \emph{Algorithms}, 15\penalty0 (6), 2022.
\newblock ISSN 1999-4893.
\newblock \doi{10.3390/a15060184}.
\newblock URL \url{https://www.mdpi.com/1999-4893/15/6/184}.

\bibitem[Amos et~al.(2024)Amos, Berant, and Gupta]{transformer_lra}
Amos, I., Berant, J., and Gupta, A.
\newblock Never train from scratch: Fair comparison of long-sequence models requires data-driven priors.
\newblock In \emph{The Twelfth International Conference on Learning Representations}, 2024.
\newblock URL \url{https://openreview.net/forum?id=PdaPky8MUn}.

\bibitem[Arjovsky et~al.(2016)Arjovsky, Shah, and Bengio]{urnn}
Arjovsky, M., Shah, A., and Bengio, Y.
\newblock Unitary evolution recurrent neural networks.
\newblock In \emph{Proceedings of the 33rd International Conference on International Conference on Machine Learning - Volume 48}, ICML'16, pp.\  1120–1128. JMLR.org, 2016.

\bibitem[Baevski \& Auli(2019)Baevski and Auli]{transformer-s4}
Baevski, A. and Auli, M.
\newblock Adaptive input representations for neural language modeling.
\newblock In \emph{International Conference on Learning Representations}, 2019.
\newblock URL \url{https://openreview.net/forum?id=ByxZX20qFQ}.

\bibitem[Bengio et~al.(1994)Bengio, Simard, and Frasconi]{bengio_vanishing}
Bengio, Y., Simard, P., and Frasconi, P.
\newblock Learning long-term dependencies with gradient descent is difficult.
\newblock \emph{IEEE Transactions on Neural Networks}, 5\penalty0 (2):\penalty0 157--166, 1994.
\newblock \doi{10.1109/72.279181}.

\bibitem[Bhattamishra et~al.(2024)Bhattamishra, Patel, Blunsom, and Kanade]{bhattamishra2023understanding}
Bhattamishra, S., Patel, A., Blunsom, P., and Kanade, V.
\newblock Understanding in-context learning in transformers and {LLM}s by learning to learn discrete functions.
\newblock In \emph{The Twelfth International Conference on Learning Representations}, 2024.
\newblock URL \url{https://openreview.net/forum?id=ekeyCgeRfC}.

\bibitem[Blelloch(1990)]{sum_prefix}
Blelloch, G.~E.
\newblock Prefix sums and their applications.
\newblock In \emph{Sythesis of parallel algorithms}, pp.\  35---60. Morgan Kaufmann Publishers Inc., 1990.
\newblock URL \url{http://citeseerx.ist.psu.edu/viewdoc/summary?doi=10.1.1.47.6430}.

\bibitem[Bradbury et~al.(2018)Bradbury, Frostig, Hawkins, Johnson, Leary, Maclaurin, Necula, Paszke, Vander{P}las, Wanderman-{M}ilne, and Zhang]{jax}
Bradbury, J., Frostig, R., Hawkins, P., Johnson, M.~J., Leary, C., Maclaurin, D., Necula, G., Paszke, A., Vander{P}las, J., Wanderman-{M}ilne, S., and Zhang, Q.
\newblock {JAX}: composable transformations of {P}ython+{N}um{P}y programs, 2018.
\newblock URL \url{http://github.com/google/jax}.

\bibitem[Burrus \& Parks(1985)Burrus and Parks]{burrus1985convolution}
Burrus, C.~S. and Parks, T.
\newblock Convolution algorithms.
\newblock \emph{Citeseer: New York, NY, USA}, 6:\penalty0 15, 1985.

\bibitem[Chen(1998)]{chenlinearsystem}
Chen, C.-T.
\newblock \emph{Linear System Theory and Design}.
\newblock Oxford University Press, Inc., USA, 3rd edition, 1998.
\newblock ISBN 0195117778.

\bibitem[Chen et~al.(2021)Chen, Lu, Rajeswaran, Lee, Grover, Laskin, Abbeel, Srinivas, and Mordatch]{decisiontransformer}
Chen, L., Lu, K., Rajeswaran, A., Lee, K., Grover, A., Laskin, M., Abbeel, P., Srinivas, A., and Mordatch, I.
\newblock Decision transformer: Reinforcement learning via sequence modeling.
\newblock In Beygelzimer, A., Dauphin, Y., Liang, P., and Vaughan, J.~W. (eds.), \emph{Advances in Neural Information Processing Systems}, 2021.
\newblock URL \url{https://openreview.net/forum?id=a7APmM4B9d}.

\bibitem[Choromanski et~al.(2021)Choromanski, Likhosherstov, Dohan, Song, Gane, Sarlos, Hawkins, Davis, Mohiuddin, Kaiser, Belanger, Colwell, and Weller]{performer}
Choromanski, K.~M., Likhosherstov, V., Dohan, D., Song, X., Gane, A., Sarlos, T., Hawkins, P., Davis, J.~Q., Mohiuddin, A., Kaiser, L., Belanger, D.~B., Colwell, L.~J., and Weller, A.
\newblock Rethinking attention with performers.
\newblock In \emph{International Conference on Learning Representations}, 2021.
\newblock URL \url{https://openreview.net/forum?id=Ua6zuk0WRH}.

\bibitem[Chung et~al.(2014)Chung, Gulcehre, Cho, and Bengio]{gru}
Chung, J., Gulcehre, C., Cho, K., and Bengio, Y.
\newblock Empirical evaluation of gated recurrent neural networks on sequence modeling.
\newblock In \emph{NIPS 2014 Workshop on Deep Learning, December 2014}, 2014.

\bibitem[Dauphin et~al.(2017)Dauphin, Fan, Auli, and Grangier]{glu}
Dauphin, Y.~N., Fan, A., Auli, M., and Grangier, D.
\newblock Language modeling with gated convolutional networks.
\newblock In \emph{Proceedings of the 34th International Conference on Machine Learning - Volume 70}, ICML'17, pp.\  933–941. JMLR.org, 2017.

\bibitem[Fu et~al.(2023)Fu, Dao, Saab, Thomas, Rudra, and R{\'e}]{h3}
Fu, D.~Y., Dao, T., Saab, K.~K., Thomas, A.~W., Rudra, A., and R{\'e}, C.
\newblock Hungry {H}ungry {H}ippos: Towards language modeling with state space models.
\newblock In \emph{International Conference on Learning Representations}, 2023.

\bibitem[Fu et~al.(2024)Fu, Kumbong, Nguyen, and R{\'e}]{flashfftconv}
Fu, D.~Y., Kumbong, H., Nguyen, E., and R{\'e}, C.
\newblock Flash{FFTC}onv: Efficient convolutions for long sequences with tensor cores.
\newblock In \emph{The Twelfth International Conference on Learning Representations}, 2024.
\newblock URL \url{https://openreview.net/forum?id=gPKTTAfYBp}.

\bibitem[Gao et~al.(2021)Gao, Biderman, Black, Golding, Hoppe, Foster, Phang, He, Thite, Nabeshima, Presser, and Leahy]{pile}
Gao, L., Biderman, S., Black, S., Golding, L., Hoppe, T., Foster, C., Phang, J., He, H., Thite, A., Nabeshima, N., Presser, S., and Leahy, C.
\newblock The pile: An 800gb dataset of diverse text for language modeling.
\newblock \emph{CoRR}, abs/2101.00027, 2021.
\newblock URL \url{https://arxiv.org/abs/2101.00027}.

\bibitem[Gao et~al.(2023)Gao, Tow, Abbasi, Biderman, Black, DiPofi, Foster, Golding, Hsu, Le~Noac'h, Li, McDonell, Muennighoff, Ociepa, Phang, Reynolds, Schoelkopf, Skowron, Sutawika, Tang, Thite, Wang, Wang, and Zou]{eval-harness}
Gao, L., Tow, J., Abbasi, B., Biderman, S., Black, S., DiPofi, A., Foster, C., Golding, L., Hsu, J., Le~Noac'h, A., Li, H., McDonell, K., Muennighoff, N., Ociepa, C., Phang, J., Reynolds, L., Schoelkopf, H., Skowron, A., Sutawika, L., Tang, E., Thite, A., Wang, B., Wang, K., and Zou, A.
\newblock A framework for few-shot language model evaluation, 12 2023.
\newblock URL \url{https://zenodo.org/records/10256836}.

\bibitem[Glorot \& Bengio(2010)Glorot and Bengio]{glorotinit}
Glorot, X. and Bengio, Y.
\newblock Understanding the difficulty of training deep feedforward neural networks.
\newblock In Teh, Y.~W. and Titterington, M. (eds.), \emph{Proceedings of the Thirteenth International Conference on Artificial Intelligence and Statistics}, volume~9 of \emph{Proceedings of Machine Learning Research}, pp.\  249--256, Chia Laguna Resort, Sardinia, Italy, 13--15 May 2010. PMLR.
\newblock URL \url{https://proceedings.mlr.press/v9/glorot10a.html}.

\bibitem[Gonzalez \& Woods(2008)Gonzalez and Woods]{gonzalez2008digital}
Gonzalez, R.~C. and Woods, R.~E.
\newblock \emph{Digital image processing}.
\newblock Prentice Hall, Upper Saddle River, N.J., 2008.
\newblock ISBN 9780131687288 013168728X 9780135052679 013505267X.

\bibitem[Gu \& Dao(2023)Gu and Dao]{mamba}
Gu, A. and Dao, T.
\newblock Mamba: Linear-time sequence modeling with selective state spaces, 2023.

\bibitem[Gu et~al.(2022{\natexlab{a}})Gu, Goel, Gupta, and R\'{e}]{s4d}
Gu, A., Goel, K., Gupta, A., and R\'{e}, C.
\newblock On the parameterization and initialization of diagonal state space models.
\newblock In Koyejo, S., Mohamed, S., Agarwal, A., Belgrave, D., Cho, K., and Oh, A. (eds.), \emph{Advances in Neural Information Processing Systems}, volume~35, pp.\  35971--35983. Curran Associates, Inc., 2022{\natexlab{a}}.

\bibitem[Gu et~al.(2022{\natexlab{b}})Gu, Goel, and Re]{s4}
Gu, A., Goel, K., and Re, C.
\newblock Efficiently modeling long sequences with structured state spaces.
\newblock In \emph{International Conference on Learning Representations}, 2022{\natexlab{b}}.
\newblock URL \url{https://openreview.net/forum?id=uYLFoz1vlAC}.

\bibitem[Gu et~al.(2023)Gu, Johnson, Timalsina, Rudra, and Re]{hippo}
Gu, A., Johnson, I., Timalsina, A., Rudra, A., and Re, C.
\newblock How to train your {HIPPO}: State space models with generalized orthogonal basis projections.
\newblock In \emph{International Conference on Learning Representations}, 2023.
\newblock URL \url{https://openreview.net/forum?id=klK17OQ3KB}.

\bibitem[Gupta et~al.(2022)Gupta, Gu, and Berant]{dss}
Gupta, A., Gu, A., and Berant, J.
\newblock Diagonal state spaces are as effective as structured state spaces.
\newblock In Koyejo, S., Mohamed, S., Agarwal, A., Belgrave, D., Cho, K., and Oh, A. (eds.), \emph{Advances in Neural Information Processing Systems 35 - 36th Conference on Neural Information Processing Systems, NeurIPS 2022}, Advances in Neural Information Processing Systems. Neural information processing systems foundation, 2022.
\newblock Publisher Copyright: {\textcopyright} 2022 Neural information processing systems foundation. All rights reserved.; 36th Conference on Neural Information Processing Systems, NeurIPS 2022 ; Conference date: 28-11-2022 Through 09-12-2022.

\bibitem[Hasani et~al.(2023)Hasani, Lechner, Wang, Chahine, Amini, and Rus]{ls4}
Hasani, R., Lechner, M., Wang, T.-H., Chahine, M., Amini, A., and Rus, D.
\newblock Liquid structural state-space models.
\newblock In \emph{The Eleventh International Conference on Learning Representations}, 2023.
\newblock URL \url{https://openreview.net/forum?id=g4OTKRKfS7R}.

\bibitem[He et~al.(2015)He, Zhang, Ren, and Sun]{heinit}
He, K., Zhang, X., Ren, S., and Sun, J.
\newblock Delving deep into rectifiers: Surpassing human-level performance on imagenet classification.
\newblock In \emph{2015 IEEE International Conference on Computer Vision (ICCV)}, pp.\  1026--1034, 2015.
\newblock \doi{10.1109/ICCV.2015.123}.

\bibitem[He et~al.(2016)He, Zhang, Ren, and Sun]{resnet}
He, K., Zhang, X., Ren, S., and Sun, J.
\newblock Deep residual learning for image recognition.
\newblock In \emph{2016 IEEE Conference on Computer Vision and Pattern Recognition (CVPR)}, pp.\  770--778, 2016.
\newblock \doi{10.1109/CVPR.2016.90}.

\bibitem[Hendrycks \& Gimpel(2023)Hendrycks and Gimpel]{gelu}
Hendrycks, D. and Gimpel, K.
\newblock Gaussian error linear units (gelus), 2023.

\bibitem[Hochreiter \& Schmidhuber(1997)Hochreiter and Schmidhuber]{lstm}
Hochreiter, S. and Schmidhuber, J.
\newblock Long short-term memory.
\newblock \emph{Neural computation}, 9\penalty0 (8):\penalty0 1735--1780, 1997.

\bibitem[Horn \& Johnson(1985)Horn and Johnson]{montel}
Horn, R.~A. and Johnson, C.~R.
\newblock \emph{Matrix Analysis}.
\newblock Cambridge University Press, 1985.

\bibitem[Katsch(2023)]{gateloop}
Katsch, T.
\newblock Gateloop: Fully data-controlled linear recurrence for sequence modeling, 2023.

\bibitem[Kidger et~al.(2020)Kidger, Morrill, Foster, and Lyons]{kidger2020neural}
Kidger, P., Morrill, J., Foster, J., and Lyons, T.
\newblock Neural controlled differential equations for irregular time series.
\newblock \emph{Advances in Neural Information Processing Systems}, 33:\penalty0 6696--6707, 2020.

\bibitem[Krizhevsky(2009)]{cifar10}
Krizhevsky, A.
\newblock Learning multiple layers of features from tiny images.
\newblock 2009.
\newblock URL \url{https://api.semanticscholar.org/CorpusID:18268744}.

\bibitem[Linsley et~al.(2018)Linsley, Kim, Veerabadran, Windolf, and Serre]{pathfinder}
Linsley, D., Kim, J., Veerabadran, V., Windolf, C., and Serre, T.
\newblock Learning long-range spatial dependencies with horizontal gated recurrent units.
\newblock In Bengio, S., Wallach, H., Larochelle, H., Grauman, K., Cesa-Bianchi, N., and Garnett, R. (eds.), \emph{Advances in Neural Information Processing Systems}, volume~31. Curran Associates, Inc., 2018.
\newblock URL \url{https://proceedings.neurips.cc/paper_files/paper/2018/file/ec8956637a99787bd197eacd77acce5e-Paper.pdf}.

\bibitem[Loshchilov \& Hutter(2019)Loshchilov and Hutter]{adamw}
Loshchilov, I. and Hutter, F.
\newblock Decoupled weight decay regularization.
\newblock In \emph{International Conference on Learning Representations}, 2019.
\newblock URL \url{https://openreview.net/forum?id=Bkg6RiCqY7}.

\bibitem[Maas et~al.(2011)Maas, Daly, Pham, Huang, Ng, and Potts]{imdb}
Maas, A.~L., Daly, R.~E., Pham, P.~T., Huang, D., Ng, A.~Y., and Potts, C.
\newblock Learning word vectors for sentiment analysis.
\newblock In Lin, D., Matsumoto, Y., and Mihalcea, R. (eds.), \emph{Proceedings of the 49th Annual Meeting of the Association for Computational Linguistics: Human Language Technologies}, pp.\  142--150, Portland, Oregon, USA, June 2011. Association for Computational Linguistics.
\newblock URL \url{https://aclanthology.org/P11-1015}.

\bibitem[Martin \& Cundy(2018)Martin and Cundy]{parallelrnn}
Martin, E. and Cundy, C.
\newblock Parallelizing linear recurrent neural nets over sequence length.
\newblock In \emph{International Conference on Learning Representations}, 2018.
\newblock URL \url{https://openreview.net/forum?id=HyUNwulC-}.

\bibitem[Massaroli et~al.(2021)Massaroli, Poli, Sonoda, Suzuki, Park, Yamashita, and Asama]{massaroli2021differentiable}
Massaroli, S., Poli, M., Sonoda, S., Suzuki, T., Park, J., Yamashita, A., and Asama, H.
\newblock Differentiable multiple shooting layers.
\newblock \emph{Advances in Neural Information Processing Systems}, 34:\penalty0 16532--16544, 2021.

\bibitem[Massaroli et~al.(2023)Massaroli, Poli, Fu, Kumbong, Parnichkun, Romero, Timalsina, McIntyre, Chen, Rudra, Zhang, Re, Ermon, and Bengio]{lhyena}
Massaroli, S., Poli, M., Fu, D.~Y., Kumbong, H., Parnichkun, R.~N., Romero, D.~W., Timalsina, A., McIntyre, Q., Chen, B., Rudra, A., Zhang, C., Re, C., Ermon, S., and Bengio, Y.
\newblock Laughing hyena distillery: Extracting compact recurrences from convolutions.
\newblock In \emph{Thirty-seventh Conference on Neural Information Processing Systems}, 2023.
\newblock URL \url{https://openreview.net/forum?id=OWELckerm6}.

\bibitem[Nangia \& Bowman(2018)Nangia and Bowman]{listops}
Nangia, N. and Bowman, S.
\newblock {L}ist{O}ps: A diagnostic dataset for latent tree learning.
\newblock In Cordeiro, S.~R., Oraby, S., Pavalanathan, U., and Rim, K. (eds.), \emph{Proceedings of the 2018 Conference of the North {A}merican Chapter of the Association for Computational Linguistics: Student Research Workshop}, pp.\  92--99, New Orleans, Louisiana, USA, June 2018. Association for Computational Linguistics.
\newblock \doi{10.18653/v1/N18-4013}.
\newblock URL \url{https://aclanthology.org/N18-4013}.

\bibitem[Oppenheim et~al.(1999)Oppenheim, Schafer, and Buck]{oppenheim99}
Oppenheim, A.~V., Schafer, R.~W., and Buck, J.~R.
\newblock \emph{Discrete-Time Signal Processing}.
\newblock Prentice-hall Englewood Cliffs, second edition, 1999.

\bibitem[Orvieto et~al.(2023)Orvieto, Smith, Gu, Fernando, Gulcehre, Pascanu, and De]{lru}
Orvieto, A., Smith, S.~L., Gu, A., Fernando, A., Gulcehre, C., Pascanu, R., and De, S.
\newblock Resurrecting recurrent neural networks for long sequences.
\newblock In \emph{Proceedings of the 40th International Conference on Machine Learning}, ICML'23. JMLR.org, 2023.

\bibitem[Pan(2001)]{structured_matrices_and_polynomials}
Pan, V.~Y.
\newblock \emph{Structured Matrices and Polynomials: Unified Superfast Algorithms}.
\newblock Springer-Verlag, Berlin, Heidelberg, 2001.
\newblock ISBN 0817642404.

\bibitem[Pascanu et~al.(2013)Pascanu, Mikolov, and Bengio]{vanishing}
Pascanu, R., Mikolov, T., and Bengio, Y.
\newblock On the difficulty of training recurrent neural networks.
\newblock In Dasgupta, S. and McAllester, D. (eds.), \emph{Proceedings of the 30th International Conference on Machine Learning}, volume~28 of \emph{Proceedings of Machine Learning Research}, pp.\  1310--1318, Atlanta, Georgia, USA, 17--19 Jun 2013. PMLR.
\newblock URL \url{https://proceedings.mlr.press/v28/pascanu13.html}.

\bibitem[Poli et~al.(2023{\natexlab{a}})Poli, Massaroli, Nguyen, Fu, Dao, Baccus, Bengio, Ermon, and R\'{e}]{hyena}
Poli, M., Massaroli, S., Nguyen, E., Fu, D.~Y., Dao, T., Baccus, S., Bengio, Y., Ermon, S., and R\'{e}, C.
\newblock Hyena hierarchy: towards larger convolutional language models.
\newblock In \emph{Proceedings of the 40th International Conference on Machine Learning}, ICML'23. JMLR.org, 2023{\natexlab{a}}.

\bibitem[Poli et~al.(2023{\natexlab{b}})Poli, Wang, Massaroli, Quesnelle, Nguyen, and Thomas]{poli2023stripedhyena}
Poli, M., Wang, J., Massaroli, S., Quesnelle, J., Nguyen, E., and Thomas, A.
\newblock Stripedhyena: Moving beyond transformers with hybrid signal processing models.
\newblock 2023{\natexlab{b}}.

\bibitem[Radev et~al.(2009)Radev, Muthukrishnan, and Qazvinian]{retrieval}
Radev, D.~R., Muthukrishnan, P., and Qazvinian, V.
\newblock The {ACL} {A}nthology network corpus.
\newblock In Kan, M.-Y. and Teufel, S. (eds.), \emph{Proceedings of the 2009 Workshop on Text and Citation Analysis for Scholarly Digital Libraries ({NLPIR}4{DL})}, pp.\  54--61, Suntec City, Singapore, August 2009. Association for Computational Linguistics.
\newblock URL \url{https://aclanthology.org/W09-3607}.

\bibitem[Radford et~al.(2018)Radford, Narasimhan, Salimans, and Sutskever]{unsupervisedllm}
Radford, A., Narasimhan, K., Salimans, T., and Sutskever, I.
\newblock Improving language understanding by generative pre-training.
\newblock 2018.

\bibitem[Ren et~al.(2023)Ren, Liu, Wang, Xu, Zhu, and Zhai]{ren2023sparse}
Ren, L., Liu, Y., Wang, S., Xu, Y., Zhu, C., and Zhai, C.
\newblock Sparse modular activation for efficient sequence modeling, 2023.

\bibitem[Romero et~al.(2022)Romero, Kuzina, Bekkers, Tomczak, and Hoogendoorn]{ckconv}
Romero, D.~W., Kuzina, A., Bekkers, E.~J., Tomczak, J.~M., and Hoogendoorn, M.
\newblock Ckconv: Continuous kernel convolution for sequential data.
\newblock In \emph{The Tenth International Conference on Learning Representations, {ICLR} 2022, Virtual Event, April 25-29, 2022}. OpenReview.net, 2022.
\newblock URL \url{https://openreview.net/forum?id=8FhxBtXSl0}.

\bibitem[Sandberg(1963)]{sandberg1963theory}
Sandberg, I.~W.
\newblock On the theory of linear multi-loop feedback systems.
\newblock \emph{Bell System Technical Journal}, 42\penalty0 (2):\penalty0 355--382, 1963.

\bibitem[Selesnick \& Burrus(2017)Selesnick and Burrus]{selesnick2017fast}
Selesnick, I.~W. and Burrus, C.~S.
\newblock Fast convolution and filtering.
\newblock In \emph{Digital Signal Processing Fundamentals}, pp.\  185--208. CRC Press, 2017.

\bibitem[Sitzmann et~al.(2020)Sitzmann, Martel, Bergman, Lindell, and Wetzstein]{siren}
Sitzmann, V., Martel, J.~N., Bergman, A.~W., Lindell, D.~B., and Wetzstein, G.
\newblock Implicit neural representations with periodic activation functions.
\newblock In \emph{Proc. NeurIPS}, 2020.

\bibitem[Smith et~al.(2023)Smith, Warrington, and Linderman]{s5}
Smith, J.~T., Warrington, A., and Linderman, S.
\newblock Simplified state space layers for sequence modeling.
\newblock In \emph{The Eleventh International Conference on Learning Representations}, 2023.
\newblock URL \url{https://openreview.net/forum?id=Ai8Hw3AXqks}.

\bibitem[Tay et~al.(2021)Tay, Dehghani, Abnar, Shen, Bahri, Pham, Rao, Yang, Ruder, and Metzler]{lra}
Tay, Y., Dehghani, M., Abnar, S., Shen, Y., Bahri, D., Pham, P., Rao, J., Yang, L., Ruder, S., and Metzler, D.
\newblock Long range arena : A benchmark for efficient transformers.
\newblock In \emph{International Conference on Learning Representations}, 2021.
\newblock URL \url{https://openreview.net/forum?id=qVyeW-grC2k}.

\bibitem[Vaswani et~al.(2017)Vaswani, Shazeer, Parmar, Uszkoreit, Jones, Gomez, Kaiser, and Polosukhin]{transformer}
Vaswani, A., Shazeer, N., Parmar, N., Uszkoreit, J., Jones, L., Gomez, A.~N., Kaiser, L.~u., and Polosukhin, I.
\newblock Attention is all you need.
\newblock In Guyon, I., Luxburg, U.~V., Bengio, S., Wallach, H., Fergus, R., Vishwanathan, S., and Garnett, R. (eds.), \emph{Advances in Neural Information Processing Systems}, volume~30. Curran Associates, Inc., 2017.
\newblock URL \url{https://proceedings.neurips.cc/paper_files/paper/2017/file/3f5ee243547dee91fbd053c1c4a845aa-Paper.pdf}.

\bibitem[Zhang et~al.(2023)Zhang, Saab, Poli, Dao, Goel, and R{\'e}]{spacetime}
Zhang, M., Saab, K., Poli, M., Dao, T., Goel, K., and R{\'e}, C.
\newblock Effectively modeling time series with simple discrete state spaces.
\newblock \emph{International Conference on Learning Representations}, 2023.

\end{thebibliography}
\bibliographystyle{icml2024}

\newpage
\appendix
\numberwithin{equation}{subsection}
\onecolumn

\rule[0pt]{\columnwidth}{1pt}
\begin{center}
    \huge{Supplementary Material}
\end{center}
\rule[0pt]{\columnwidth}{1.5pt}

\tableofcontents

\vspace{1em}
{\large \bf Author Contribution}

\begin{itemize}
    \item[] \textbf{R.N.P.} Developed the algorithm, theory, code base, and manuscript. Managed and conducted experiments.
    \item[] \textbf{S.M.} Developed the algorithm, theory, and manuscript. Supervised research.
    \item[] \textbf{A.M.} Developed the code base and manuscript. Conducted experiments and secured compute.
    \item[] \textbf{J.S.} Reviewed manuscript and assisted in writing.
    \item[] \textbf{R.H.}, \textbf{M.L.} Reviewed manuscript and secured compute.
    \item[] \textbf{Q.A.}, \textbf{C.R.}, \textbf{H.A.}, \textbf{S.E.}, \textbf{T.S.} Supervised research.
    \item [] \textbf{A.Y.} Supervised research and secured compute.
    \item[] \textbf{M.P.} Developed the algorithm, theory and manuscript. Supervised research.
\end{itemize}

\newpage
\section{Linear System Theory} \label{sec:linsys}

This section delves into linear system theory, including denomination of various characteristics such as lumpedness, time-invariance, etc., and also includes the analysis and derivation of $\cZ$-domain transfer functions.

\subsection{Overview and Basics}

\paragraph{Linear Systems:} Linear systems consist of a series of linear equations generally expressed as:
\begin{equation}
y = \sG u,
\end{equation}
in which $u \in \mathbb{R}^{\ell}$, $y \in \mathbb{R}^{\ell}$, and $\sG \in \mathbb{R}^{T \times T}$ are the input, output, and the transformation matrix, respectively. These systems adhere to the principles of linearity, including additivity and homogeneity. For the purpose of processing sequences, they can also written as:
\begin{equation}\label{eq:lin_sys}
y_t = \sum_{j = t_0}^{t} \sG_{t, t-j}u_j,
\end{equation}
in which, $\sG_{t, t-j}$ scales the input signal $u_j$ for the output, based on the absolute time $t$ and the relative time $t-j$.

\paragraph{Time-Invariance} A linear time-invariant (LTI) system simply discards the absolute time dependence in (\ref{eq:lin_sys}) as follows:
\begin{equation}
    y_t = \sum_{j = t_0}^{t} h_{t-j}u_j.
\end{equation}
These systems are equivalent to convolutions characterized by $h$, with a shorthand notation $y_t = (h \ast u)_t$. 
$h$ is also known as the system's \textit{impulse response}. As $y = h \ast \delta = h$, in which $\delta$ is the Kronecker delta (impulse) function. 

\paragraph{Lumped Systems:} Lumped LTI systems \cite{chenlinearsystem} are LTI systems that can be characterized with a finite and discrete (lumped) set of states. They can be formulated as a state-space model:
\begin{equation}
    \begin{aligned}
        x_{t+1} &= \sA x_t + \sB u_t\\
        y_t &= \sC x_t + h_0 u_t,
    \end{aligned}
\end{equation}
where $\sA \in \bC^{N \times N}$, $\sB \in \bC^{N \times 1}$, $\sC \in \bC^{1 \times N}$, and $h_0 \in \R$. Unrolling the recurrence, its connection to the convolutional operation could be made clear:
\begin{equation}
    \begin{aligned}
        y_0 &= \sC x_0 + h_0 u_0 \\
        y_1 &= \sC (\sA x_0 + \sB u_0) + h_0 u_1\\
        y_2 &= \sC (\sA (\sA x_0 + \sB u_0) + \sB u_1) + h_0 u_2\\
        \vdots\\
        y_t &= h_0 u_t + \sum_{j= 1}^{t}{\sC \sA^{j - 1}\sB u_{t-j}} + \sC\sA^{t} x_0 \\
        y_t &= (h\ast u)_t + \sC\sA^t x_0, \; \text{where } h_t = \begin{cases}
            h_0 & t = 0\\
            \sC\sA^{t-1}\sB & t > 0
        \end{cases}.
    \end{aligned}
\end{equation}
Note that all lumped LTI systems have complex exponential convolutional kernels. Non-lumped systems are not restricted to exponential convolutional kernels but cannot be directly expressed using a fixed and finite state-space, i.e. they have a non-constant time autoregressive inference complexity. Convolutional filters implicitly parameterized by MLPs such as CKConv \cite{ckconv} and \cite{hyena} are examples of non-lumped linear time-invariant systems.

\subsection{Transfer Function Realization of Lumped LTI Systems} \label{sec:tf_realization}

\paragraph{Control Theorists Derivation:} By applying the shift forward operator ($z$) in $\cZ$-domain to the state-space equations, we can obtain its \textit{transfer function} as follows.
\begin{equation}
    \begin{aligned}
        x_{k+1} &= \sA x_k + \sB u_k && \text{state dynamics}\\
        X(z)z &= \sA X(z) + \sB U(z) && \text{$\cZ$-transform}\\
        (z\sI - \sA)X(z) &= \sB U(z) && (z\sI-\sA)\text{ is also known as the resolvent matrix}\\
        X(z) &= (Iz - \sA)^{-1}\sB U(z) \quad \\
        H(z) &= \frac{Y(z)}{U(z)} = \sC(z\sI - \sA)^{-1}\sB + h_0 \quad &&  \text{substituted $X(z)$ into $Y(z) = \sC X(z) + h_0 U(z)$}
    \end{aligned}
\end{equation}

\paragraph{Alternative Derivation: \cite{lhyena}} The transfer function can also be derived by direct $\cZ$-transform of the impulse response $h_t$ of the system. This derivation is useful to highlight the region of convergence of the transfer function.
\begin{equation}\label{eq:impulse_response_tf}
	\begin{aligned}
		H(z) &= h_0 + \sum_{t=1}^\infty z^{-t} \sC \sA^{t-1} \sB  && \text{$h_0$ is pulled out via $h_0 z^0=h_0$}\\
		&= h_0 + \sC\left[\sum_{t=1}^\infty z^{-t} \sA^{t-1} \right]\sB && \text{multiplication distributes over sum.}\\
		&= h_0 + z^{-1}\sC\left[\sum_{t=1}^\infty z^{-(t-1)}\sA^{t-1}\right]\sB  && \text{multiply by $z/z$}\\
        &= h_0 + z^{-1}\sC\left[\sum_{t=0}^\infty (z^{-1}\sA )^t\right]\sB && \text{change of index and collect like terms}\\
	\end{aligned}
\end{equation}
We look at the convergence of the series $\sum_{t=0}^\infty \|z^{-1}\sA \|^t_2$. We have 
\begin{equation*}
    \begin{aligned}
        \|z^{-1}\sA \|_2 &\leq \|z^{-1}\|_2\|\sA\|_2\\
        &= \|r^{-1}e^{-i\omega}\|_2\|\sA\|_2 && \text{using $z\coloneqq re^{i\w}\in\bC,~r,\w\in\R$}\\
        &\leq r^{-1}\|\sA\|_2 = r^{-1}\rho(\sA)
    \end{aligned}
\end{equation*}
The series converges to $1 / (1 - r^{-1}\rho(\sA))$ if and only if $r^{-1}\rho(\sA)<1$ i.e. for $r>\rho(\sA)$. Thus, in the exterior of the disk with radius $\rho(\sA)$, $\bD_{\rho(\sA)} \coloneqq \{z\in\bC : |z|>\rho(\sA)\}$, $\sum_{t=0}^\infty (z^{-1}\sA)^t$ converges to $(\sI - z^{-1}\sA)^{-1}$ and 
\begin{equation*}
    z\in\bD_{\rho(\sA)}~\Rightarrow~H(z) = h_0 + z^{-1}\sC(\sI - z^{-1}\sA)^{-1}\sB =  h_0 + \sC(z\sI - \sA)^{-1}\sB
\end{equation*}
The transfer function $H(z)=h_0 +\sC(z\sI -A)^{-1}\sB$ of a stable lumped discrete-time system is defined outside the disc in the complex plane that encloses all the eigenvalues of $\sA$.

Further dissecting $H(z)=h_0 +\sC(z\sI -A)^{-1}\sB$, note that to compute the inverse, $\det(z\sI - \sA)$ is a $n$th order Monic polynomial, and $\sC[\adj(z\sI - \sA)]\sB$ is a $n-1$ order polynomial (for the SISO case), hence the general form of a transfer function can be written in the form of the following rational function (this is discussed in greater detail in \ref{sec:ssm2tf}):
\begin{equation}\label{eq:tf_coef}
    H(z) = \frac{b_{1}z^{-1} + b_2 z^{-2} + \dots + b_n z^{-n}}{1 + a_{1}z^{-1} + a_2 z^{-2} + \dots + a_n z^{-n}}+ h_0 \quad \rightarrow \text{Rational function form}.
\end{equation}
The SISO rational coefficient form has $2n+1$ parameters. With partial fraction decomposition, the rational function can be broken down into its first order partial decomposition, resulting in a modal representation:
\begin{equation}\label{eq:tf_modal}
    H(z) = \sum_{i = 1}^{n}{\frac{r_i}{z-\lambda_i}} + h_0 \quad \rightarrow \text{Modal form},
\end{equation}
in which $r,\lambda \in \bC$. This form parameterizes the \textit{poles} ($\lambda$) and its associated magnitude ($r$). The modal form has $2n+1$ trainable parameters. It is worth noting that the first order partial fraction decomposition does not permit any form of repeated roots, for this reason, it is not a complete representation of a lumped LTI systems.

Another way in which rational functions can be structured is called the zero-pole-gain (ZPK) representation:
\begin{equation}\label{eq:tf_zpk}
    H(z) = k\frac{\prod_{i=1}^{n-1}(z-z_i)}{\prod_{i = 1}^{n}(z-\lambda_i)} + h_0 \quad \rightarrow \text{Zero-Pole-Gain form},
\end{equation}
in which, $k$, $z$, and $\lambda$ are the gain, zeros, and poles respectively. The ZPK form has $2n+1$ trainable parameters.

\subsection{From State-Space to Transfer Function \cite{lhyena}}\label{sec:ssm2tf}
We detail an implementation oriented method to compute the coefficients $(a_i)_{i=1}^n,(b_i)_{i=1}^n$ of a SSM's transfer function. Expanding the inverse of the resolvent matrix, recall that
\begin{equation}\label{eq:tf_simple}
    H(z) = \sC[z\sI - \sA]^{-1}\sB + h_0 = \frac{\sC\adj(z\sI - \sA)\sB + {\sf det}(z\sI - \sA)h_0}{{\sf det}(z\sI - \sA)}
\end{equation}
This shows that the denominator coefficients $(a_i)_{i=1}^n$ are simply the coefficients of the characteristic polynomial of matrix $\sA$. They can be easily obtained by 1. computing the eigenvalues of $\sA$ and 2. calculating the coefficients of the polynomial whose roots are such eigenvalues. On the other hand, the numerator apparently involves more complex symbolic manipulation. This can be simplified recalling a classic matrix-determinant identity:
\begin{tcolorbox}[enhanced, frame hidden, sharp corners, boxsep=0pt, before skip=0pt, after skip=0pt, colback=ghostwhite]
\begin{lemma}[\cite{sandberg1963theory}]\label{lemma:mat_det_id}
    Let $\sM$, $\sB$, and $\sC$ respectively denote matrices of orders $n\x n$, $n\x 1$, and $1\x n$. Then,
    \begin{equation*}
        {\sf det}(\sM + \sB\sC) = {\sf det}(\sM) + \sC \adj(\sM)\sB.   
    \end{equation*}
\end{lemma}
\end{tcolorbox}
Applying Lemma~\ref{lemma:mat_det_id} to \eqref{eq:tf_simple} we obtain 
\begin{equation*}
    H(z) = \frac{{\sf det}(z\sI - \sA + \sB\sC) + {\sf det}(z\sI - \sA)(h_0 - 1)}{{\sf det}(z\sI - \sA)}.
\end{equation*}
Let ${\sf poly}(r)$ denote the coefficients of the polynomials with roots $r=(r_1,\dots,r_n)$. Then $a = {\sf poly}({\sf eig}(\sA))$.
Since $\sA$ and $\sA - \sB\sC$ are of equal dimension, their characteristic polynomials have equal order and therefore
\begin{equation*}
    b = {\sf poly}({\sf eig}(\sA - \sB\sC)) + {\sf poly}({\sf eig}(\sA))(h_0 - 1)
\end{equation*}
\begin{lstlisting}[language=python, caption={State-space $\to$ transfer function conversion code}, label={lst:recurrence}]
def get_tf_from_ss(A,B,C,h0):
    a = poly(eig(A))
    b = poly(eig(A - outer(B,C))) + (h0-1)*a
    return a, b
\end{lstlisting} 

\subsection{From Transfer Function to State-Space \cite{lhyena}}\label{sec:tf2ssm}
\paragraph{Chen's derivation}
The derivation is based on the steps reported for the continuous-time \textit{multi-input multi-output} case in \cite{chenlinearsystem} adapted to single-input single-output Transfer Functions. 

Let $H(z) = \frac{q(z)}{p(z)} + h_0$, we define a pseudo-state $v$ such that 
\begin{equation}\label{eq:chen_pseudo_state}
	p(z)V(z) = U(z)\quad \Leftrightarrow \quad V(z) = \frac{1}{p(z)}U(z).
\end{equation}
Then, we define the state $x_t\coloneqq(x_t^{1},\dots,x_t^{n})\in\R^{n}$ as 
\begin{equation}\label{eq:chen_state}
	x_t = (v_{t-1}, v_{t-2}, \cdots, v_{t-n}) \quad \Leftrightarrow \quad \cZ\{x\}(z) = X(z) = \begin{bmatrix}
		z^{-1} \\
		\vdots \\
		z^{-n}
	\end{bmatrix}V(z).
\end{equation}
From \eqref{eq:chen_pseudo_state} we have
\begin{equation*}
	\begin{aligned}
		V(z) + a_1z^{-1}V(z) + \cdots + a_nz^{-n}V(z) = U(z) &~~ \Leftrightarrow\\
		V(z) = -a_1z^{-1}V(z) - \cdots - a_nz^{-n}V(z) + U(z) &~~ \Leftrightarrow\\
		v_t = -a_1v_{t-1} - \cdots - a_nv_{t-n} + u_t &~~ \Leftrightarrow && \text{time-delay prop. of $\cZ$-transform}\\
		x^1_{t+1} = -a_1x^1_t - \cdots - a_nx^n_t + u_t &~~ \Leftrightarrow && \text{by def. of state \eqref{eq:chen_state}}.\\
	\end{aligned}
\end{equation*}
Thus, we have the overall recurrence
\begin{equation*}
	\begin{aligned}
		x^1_{t+1} &= -a_1x^1_t - \cdots - a_nx^n_t + u_t \\
		x^2_{t+1} &= x^1_t \\
		&\vdots \\
		x^n_{t+1} &= x^{n-1}_t \\
	\end{aligned}
\end{equation*}
which can be written in matrix form as
\begin{equation*}
	\begin{aligned}
		x_{t+1} &=
		\begin{bmatrix}
			-a_1 & -a_2 & \cdots & -a_n \\
			1 & 0 & \cdots & 0 \\
			0 & 1 & \cdots & 0 \\
			\vdots & \vdots & \ddots & \vdots \\
			0 & 0 & \cdots & 0
		\end{bmatrix}
		x_t + 
		\begin{bmatrix}
			1 \\
			0 \\
			0\\
			\vdots \\
			0
		\end{bmatrix} u_t
	\end{aligned}
\end{equation*}
The output spectrum is then given by 
\begin{equation*}
	\begin{aligned}
		Y(z) &= H(z) U(z) =  \frac{q(z)}{p(z)}U(z) + h_0U(z)\\
				&= q(z)V(z) + h_0U(z) &&\text{by def. of $V(z)$}.
	\end{aligned}	
\end{equation*}
Therefore,
\begin{equation*}
	\begin{aligned}
		Y(z) &= q(z)V(z) + h_0U(z) = \begin{bmatrix}
			b_1 & b_2 & \cdots & b_N
		\end{bmatrix}
		\begin{bmatrix}
			z^{-1} \\
			z^{-2} \\
			\vdots \\
			z^{-n}
		\end{bmatrix}V(z)+ h_0U(z)\\
		& = \begin{bmatrix}
			b_1 & b_2 & \cdots & b_n
		\end{bmatrix}X(z) + h_0U(z)
	\end{aligned}
\end{equation*}
and the output equation in time-domain is given by
\begin{equation*}
	y_t = \begin{bmatrix}
		b_1 & b_2 & \cdots & b_n
	\end{bmatrix}x_t + h_0u_t.
\end{equation*}
yielding state-space matrices \eqref{eq:ssm_canon}.
\begin{equation}\label{eq:ssm_canon}
    \left[
        \begin{array}{c|c}
        \sA & \sB\\
        \hline
        \sC & h_0
        \end{array} 
    \right]
    =
    \left[
        \begin{array}{c|c}
            \begin{matrix}
                -a_1 & -a_2 &\cdots & -a_{n-1} &- a_n \\
                1 & 0 & \cdots & 0 & 0 \\
                0 & 1 & \cdots & 0 & 0 \\
                \vdots & \vdots & \ddots & \vdots & \vdots\\
                0 & 0 & \cdots& 1 & 0
            \end{matrix} & \begin{matrix}1 \\ 0 \\ 0 \\ \vdots \\ 0\end{matrix}\\
            \hline
            \begin{matrix}~~b_1 & ~~b_2 & ~\cdots~ & ~~b_{n-1} & ~~b_n\end{matrix} & h_0
        \end{array} 
    \right] .
\end{equation}

\newpage
\section{RTF: Further Details}
\subsection{Fast Companion Recurrence} \label{app:companion_recurrence}

The recurrent step of a generic SSM \eqref{eq:dtssm} with dense system matrices usually requires $\cO(n^2)$ operations due to the matrix-vector product $\sA x_t$. We show how the recurrence of SSMs in \textit{companion canonical form}, i.e. with system's matrices \eqref{eq:ssm_canon}, requires only $\cO(n)$ operations. 
\begin{tcolorbox}[enhanced, frame hidden, sharp corners, boxsep=0pt, before skip=0pt, after skip=0pt, colback=ghostwhite]
\begin{lemma}\label{lemma:efficient_recurrence}
    The recurrent step of a state-space model in companion canonical form \eqref{eq:ssm_canon} can be evaluated in $\cO(n)$ time and memory. 
\end{lemma}
\end{tcolorbox}
\proof
The companion state matrix $\sA$ can be broken down into a lower shift matrix $\sL_n$ and a low-rank term. Particularly, with $e_1$ the first element of the canonical basis of $\R^n$ and $a = (a_1, \ldots, a_n)$, we have
\begin{equation*}
	\sA = \sL_n - e_1\otimes a.
\end{equation*}
It follows that the recurrent update can be simplified to
\begin{equation*}
	\begin{aligned}
		x_{t+1} &= \left(\sL_n - e_1\otimes a\right) x_t + \sB u_t\\
		y_t &=\sC x_t + h_0u_t
	\end{aligned}
\end{equation*}
The peculiarity of this formulation is that we never need to construct the full transition matrix to perform the recurrence. In particular we have:
\begin{equation*}
	\begin{aligned}
		x^1_{t+1} &= u_t - a^\top x_t\\
		x^{2:n}_{t+1} &= {\sf shift}(x_t)\\ 
		y_t &= b^\top x_t + h_0u_t 
	\end{aligned}
\end{equation*}
Thus, each step only requires two inner products ($n$ multiplications and $n$ sums each) and one shift operation, totaling $\cO(n)$ operations. 
\endproof

\subsection{Initialization and Stability}

Initialization schemes can significantly impact the performance of SSMs, as explored in \cite{s4}, \cite{hippo}, \cite{lru}, and \cite{spacetime}.

Intriguingly, rational transfer functions allow for initialization schemes that can be directly translated from explicitly parameterized convolutional kernels, as demonstrated below:
\begin{equation}
K_{\sf{FIR}}(z) = k_0 + k_1 z^{-1} + k_2 z^{-2} + \dots + k_{m-1} z^{-(m-1)}
\end{equation}
where $K_{\sf{FIR}}$ represents the z-domain representation of an $m$-length finite impulse response (a convolutional kernel of size $m$). It could easily be seen that by simply setting $h_0= k_0$, $a_i = 0$ and $b_i = k_{i}$ for $i \in [m]$, the rational transfer function would represent the convolutional kernel.
This implies that initialization approaches developed for explicitly parameterized convolutional models, such as \cite{heinit} and \cite{glorotinit}, can be directly applied to the rational transfer function representation.

Besides initialization, it is generally desirable for SSMs to be stable, meaning that the roots of the rational transfer function denominator (poles) should reside within a complex unit circle in the z-domain \cite{chenlinearsystem}. When employing a polar representation of the kernel eigenvalues (poles), in which the roots are parameterized by $\lambda = r e^{i\theta}$, the roots $r$ can easily be restricted to $|r| \leq 1$ in various ways such as $r = \mathsf{exp}(-\mathsf{exp}(\nu))$, where $\nu \in \mathbb{R}^{n}$ as described in \cite{lru}. However, for rational transfer functions, where the denominator is represented as a polynomial, ensuring the stability of the SSM is more challenging. \citeauthor{constrain_roots} presents several methods for constraining polynomial coefficients, for their roots to lay within the complex unit circle. One such method, Montel's method \cite{montel}, constrains the polynomial roots as follows:
\begin{equation}
\sum_{i=0}^{n-1}{|\alpha_i|} \leq 1,
\end{equation}
This can be implemented straightforwardly using a softmax or an $l_1$ norm over $n+1$ parameters, and then selecting $n$ parameters from this set, as shown in the following code snippet:
\begin{lstlisting}[language=Python]
def get_constrained_coefs(coefs_plus_scalar):
    """
    coefs_plus_scalar: torch.Tensor of shape [n+1]
    """
    return (coefs/sum(coefs.abs()))[:n] # returns n coefficients that are constrained according to Montel's method.
\end{lstlisting}\label{algo:l1_montel}
Spacetime \cite{spacetime} also utilizes this approach to bound the gradients of their SSMs during training. However, we have found that Montel's method could excessively constrain the SSMs, potentially leading to diminished performance, as shown in Table \ref{table:wt103-constrain}.

\begin{table}[h]
    \small
    \centering
    \setlength{\tabcolsep}{4pt}
    \caption{
    An ablation of different initialization and parameter constraining approaches. }
    \begin{tabular}{@{}l|cc@{}}
    \toprule
    Model & Wikitext-103 (25 epochs) & LRA Image \\
    & ppl. $\downarrow$ & acc. $\uparrow$\\
    \midrule 
    RTF + Xavier Init. + Montel Constraint & 26.512 & 89.2\\
    RTF + Xavier. Init. & - & 90.0\\
    RTF + Impulse Init. & \textbf{26.093} & \textbf{90.1}\\
    \bottomrule    
    \end{tabular}
    \label{table:wt103-constrain}
\end{table}

Next, we use a 2nd order polynomial case, as a visual illustration of the over-constraining occurring with Montel's method over the parameter space.
Given a polynomial $z^2 + \alpha_1 z + \alpha_0$, its roots can be analytically computed with:
\begin{equation}
    r = \frac{-\alpha_1 \pm \sqrt{\alpha_1^2-4\alpha_0}}{2}.
\end{equation}

In the case that $\alpha_1^2 - 4\alpha_0 < 0$, the quadratic equation becomes a summation of a real term and an imaginary term, therefore we can constrain the root to be within the unit circle by computing its norm as follows:
\begin{align}
\sqrt{\left(\frac{\alpha_1}{2}\right)^2 - \left(\frac{\sqrt{\alpha_1^2-4\alpha_0}}{2}\right)^2} &\leq 1 \\
\frac{\alpha_1^2-\alpha_1^2+4\alpha_0}{4} &\leq 1 \\
\alpha_0 \leq 1.
\end{align}

\begin{figure}[h]
    \centering
    \includegraphics[trim={-10mm 3mm 0 0},clip, width=.7\textwidth]{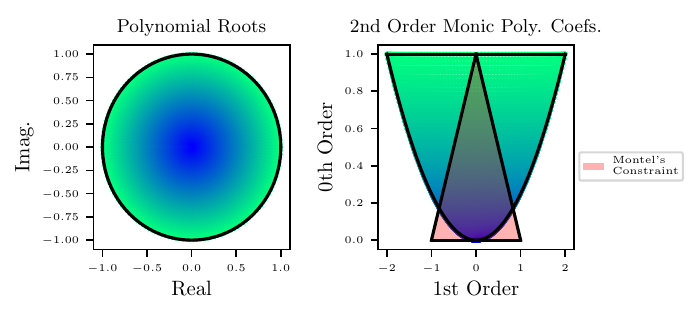}
    \caption{The space of stable roots of a 2nd order polynomial with conjugate roots is illustrated with a green-blue colormap. The figure on the right overlays the space of coefficients that obey Montel's constraints in pink.}
    \label{fig:roots}
\end{figure}

This shows that the two equations that govern the possible stable regions (for pairs of conjugate roots) are, $\alpha_0 \leq 1$ and $\alpha_0 > \frac{1}{4}\alpha_1^2$. Figure \ref{fig:roots} illustrate the space of stable coefficients with a green-blue colormap along with the space of coefficients that obey Montel's constraints in pink. Notice that a sizable portion of the coefficient space that represents a stable SSM with low decay rates is not accessible with the constraint, which hurts SpaceTime's expressivity and enforces a short term bias to the model.

We observed empirically (i.e., Table \ref{table:wt103-constrain}) that setting both numerator and denominator parameters to zeros, and setting $h_0=1$, as formulated below,
\begin{equation}
H_{\delta}(z) = 1 + \frac{0}{z^{n}},
\end{equation}
generally resulted in RTF having faster training convergence, while simultaneously avoiding instability issues that may be caused via other initialization schemes.
The improved stability of this initialization scheme is likely due to it being optimal with respect to satisfying the Montel constraint as follows:
\begin{equation}
\argmin_{\mathbf{\alpha}}(\sum_{i=0}^{n-1}{|\alpha_i|}) = \mathbf{0}.
\end{equation}
We denote this as the \textbf{zero initialization} scheme, and use it throughout all our experiments unless stated otherwise.
\newpage 
\subsection{Alternative Inference Algorithms}

\subsubsection{RTF Kernel Generation via Long Polynomial Division}

Given a rational transfer function (TF) representing an infinite length convolutional kernel:
\begin{equation}
      H(z) = h_0 + \frac{N(z)}{D(z)} = h_0 + \frac{\sum^{n-1}_0{b_i z^i}}{\sum^{n}_0{a_i z^i}} = h_0 + h_1 z^{-1} + h_2 z^{-2} + \dots,
\end{equation}
we would like to directly obtain the truncated (finite length) representation of such a kernel, in order to 1. train RTF numerators that directly correspond to the recurrent form without the need to correct for truncation (which could offer significant speedups in online learning tasks such as reinforcement learning), 2. directly evaluate the truncated transfer function $H(z)$ at $2\ell$ points, avoiding the need to convert the frequency domain kernel into time domain for causal padding.

We could take the approach of constructing an infinite length \textit{tail} function, which upon being subtracted from the original TF, results in truncation as follows:
\begin{equation}
    H_\ell(z) = H(z) - \tilde{H}_\ell(z) = h_0 + h_1 z^{-1} + h_2 z^{-2} + \dots + h_{\ell-1} z^{-\ell+1}. \label{eq:get_fir_tf}
\end{equation}
To satisfy such an equation, we observe that $\tilde{H}_\ell(z) = h_{\ell}z^{-\ell} + h_{\ell+1}z^{-(\ell+1)} + \dots$, which could be obtained from the original rational transfer function via long division of $N(z) z^{L}$ against $D(z)$ as shown below:
\begin{align}
    \frac{N(z) z^{\ell-1}}{D(z)} &= \underbrace{h_0 z^{\ell-1} + h_1 z^{\ell-2} + h_2 z^{\ell-3} + \dots + h_{\ell-1}}_{C(z)} + \underbrace{h_{\ell}z^{-1} + h_{\ell+1}z^{-2} + \dots}_{\tilde{H}_\ell(z)z^{-\ell+1}}\\
    &= C(z) + \tilde{H}_{\ell}(z) z^{-\ell+1} = C(z) + \frac{R(z)}{D(z)},
\end{align}
\begin{equation}
    \tilde{H}_{\ell}(z) = \frac{R(z)}{D(z)z^{\ell-1}}.
\end{equation}
The naive long division algorithm takes $2np$ operations, in which $p=\ell-n+1$, however with fast Toeplitz matrix inversion algorithms described in \cite{structured_matrices_and_polynomials}, such an algorithm could operate with complexity of $\cO(\ell \log \ell)$, assuming $n \ll \ell$.

Next, by simply constructing the truncated transfer function $H_\ell(z)$ via Equation (\ref{eq:get_fir_tf}), the padded convolutional kernel in frequency domain can be obtained via transfer function evaluation at $2\ell$ points of unity.

\subsubsection{Multi-Input Multi-Output RTF} \label{app:mimortf}

A multi-input multi-output (MIMO) LTI SSM could be represented using a $d \times d$ matrix of numerator polynomials, that shares a denominator polynomial, forming a rational function for each input to output pair. \citeauthor{chenlinearsystem} shows that such a system could be converted back into an SSM realizing the companion form \eqref{eq:companion_ssm} as follows:
\begin{equation}
\begin{aligned}
x_{k+1} &= \begin{bmatrix}
    -a_0 \sI_d & -a_1 \sI_d & \dots & -a_{n-2} \sI_d & -a_{n-1} \sI_d \\
    \sI_d & \mathsf{0} & \dots & \mathsf{0} & \mathsf{0} \\
    \mathsf{0} & \sI_d & \dots & \mathsf{0} & \mathsf{0} \\
    \vdots & \vdots & & \vdots & \vdots \\
    \mathsf{0} & \mathsf{0} & \dots & \sI_d & \mathsf{0}
\end{bmatrix}x_k + \begin{bmatrix}
    \sI_d \\ 
    \mathsf{0} \\
    \mathsf{0} \\
    \vdots \\
    \mathsf{0}
\end{bmatrix}u  \\
y &= \sC x_k + \sD u, 
\end{aligned}
\end{equation}

in which $\sI_d$ is a rank $d$ identity matrix, $\sC \in \mathbb{R}^{d \times nd}$ corresponds to the matrix of numerator coefficients and $\sD \in \mathbb{R}^{d \times d}$. $a_i$ is the denominator polynomial coefficient at order $i$. We can observe that such a system's $\sC$ matrix becomes excessively large, making it not competitive in terms of both parallel inference and autoregressive inference speeds against other MIMO systems. For this reason, we focus on the multi SISO \eqref{eq:multisiso_hyena} companion realization, in which the SSMs are independent across the channel dimension, channel mixing is only done afterwards with a linear projection.

\newpage
\section{Experiments}
\subsection{Memory and Latency Profiling Experiments} \label{app:memory}

\begin{figure}[h]
    \centering
    \includegraphics[trim={0 3mm 0 0},clip, width=.45\textwidth]{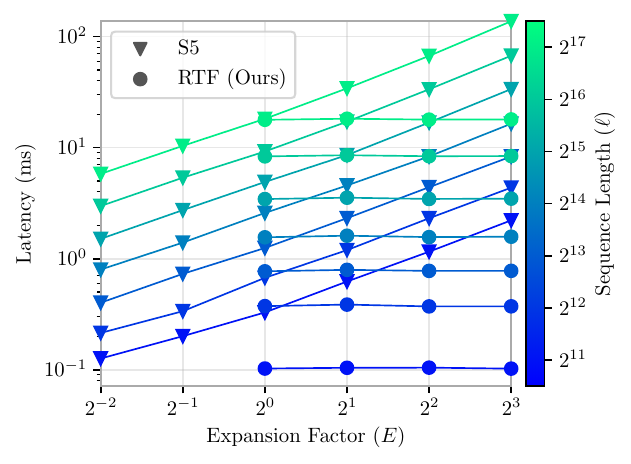}
    \caption{This figure illustrates the scaling of parallel inference latency on S5 and RTF across various sequence lengths and state sizes. When comparing equal expansion factors, it becomes evident that RTF provides lower latencies across different sequence lengths.}
    \label{fig:parallel_inference_latency}
\end{figure}

\begin{itemize}
    \item Experiments were conducted using JAX \cite{jax} on a single A100 80GB GPU for the memory profiling experiments, and on a single H100 80GB GPU for the latency profiling experiments.
    \item S5 implementation was taken directly from \cite{s5}.
    \item The memory profiling was done on a single SSM layer with channel size $d = 1024$, whereas the latency profiling was done using $d=128$.
    \item Due to S5 being a Multi-Input Multi-Output (MIMO) SSM and RTF being a Single-Input Single-Output SSM, there are few additional points to note on interpreting the results:
    \begin{itemize}
        \item For fairness we considered a RTF layer with channel mixing, which includes an additional output linear projection layer that mixes the channel dimensions.
        \item The RTF layer with channel mixing is equivalent to a block diagonal MIMO SSM with a combined state size of $n_{M} = dn$. Mamba \cite{mamba} makes use of the term \textit{state expansion factor} ($e$), which describes the state size per channel. For a multi-SISO SSM such as RTF, $e = n$, whereas for a MIMO SSM such as S5, $e = n/d$.
        \item Figure \ref{fig:mem} and Table \ref{table:memory_vals} compare each SSM layer's memory usage across multiple state sizes ($n$), whereas Figure \ref{fig:parallel_inference_latency} and Table \ref{table:latency_vals} compare SSM layer's parallel inference latency across multiple the expansion factors ($e$).
    \end{itemize}
    \item For each sequence length $\ell$, we collected profiling speeds for RTF and S5 with state-sizes ranging from 256 up to $\ell/2$.
    \item Table \ref{table:memory_vals} lists the exact peak memory usage in MB. Runs which ran out of the 80GB GPU memory is denoted as \texttt{OOM} (Out Of Memory).
    \item Figure \ref{fig:parallel_inference_latency} and Table \ref{table:latency_vals} illustrates the median parallel inference latencies (across 100 iterations) in milliseconds. 
\end{itemize}

\begin{table}[h]
    \centering
    \setlength{\tabcolsep}{4pt}
    \caption{Comparison of peak memory usage of S5 and RTF across different state-sizes and sequence lengths in MB.}
    \begin{tabular}{|l|l|c|c|c|c|c|c|}
    \hline
    \diagbox{$n$}{$\ell$}& Model & $2^{12}$ & $2^{13}$ & $2^{14}$ & $2^{15}$ & $2^{16}$ & $2^{17}$ \\
    \hline
    \multirow{2}{*}{$2^8$} & S5
    & 178.0 & 266.0 & 459.0 & 794.01 & 1510.0 & 3010.0 \\
    \cline{2-8} & RTF & 230.0 & 454.04 & 902.0 & 1760.0 & 3510.0 & 7010.0\\
    \hline
    \multirow{2}{*}{$2^9$} & S5 & 192.0 & 288.01 & 480.0 & 864.01 & 1590.0 & 3090.0 \\
    \cline{2-8} & RTF & 232.04 & 456.04 & 904.0 & 1760.0 & 3510.0 & 7010.0\\
    \hline
    \multirow{2}{*}{$2^{10}$} & S5 & 220.0 & 332.0 & 592.0 & 1140.0 & 2140.0 & 4270.0\\
    \cline{2-8} & RTF & 236.0 & 460.04 & 908.0 & 1760.0 & 3510.0 & 7010.0\\
    \hline
    \multirow{2}{*}{$2^{11}$} & S5 & 308.0 & 544.02 & 1030.0 & 1780.0 & 3530.0 & 7030.0\\
    \cline{2-8} & RTF & 244.0 & 468.04 & 916.0 & 1770.0 & 3520.0 & 7020.0\\
    \hline
    \multirow{2}{*}{$2^{12}$} & S5 & - & 960.04 & 1690.0 & 3310.0 & 6560.0 & 13060.0\\
    \cline{2-8} & RTF & - & 484.04 & 932.0 & 1790.0 & 3540.0 & 7040.0\\
    \hline
    \multirow{2}{*}{$2^{13}$} & S5 & - & - & 3130.0 & 6130.0 & 12130.0 & 24130.0\\
    \cline{2-8} & RTF & - & - & 964.0 & 1820.0 & 3570.0 & 7070.0\\
    \hline
    \multirow{2}{*}{$2^{14}$} & S5 & - & - & - & 11750.0 & 23250.0 & 46250.0\\
    \cline{2-8} & RTF& - & - & - & 1880.0 & 3630.0 & 7130.0\\
    \hline
    \multirow{2}{*}{$2^{15}$} & S5 & - & - & - & - & 49500.0 & \texttt{OOM} \\
    \cline{2-8} & RTF & - & - & - & - & 3750.0 & 7250.0 \\
    \hline
    \multirow{2}{*}{$2^{16}$} & S5 & - & - & - & - & - & \texttt{OOM} \\
    \cline{2-8} & RTF & - & - & - & - & - & 7500.0 \\
    \hline
    \end{tabular}
    \label{table:memory_vals}
\end{table}
\begin{table}[h]
    \centering
    \setlength{\tabcolsep}{4pt}
    \caption{Comparison of parallel inference latency of a single SSM layer in milliseconds across different sequence lengths and expansion factors ($e$) of RTF and S5. We report the median value across 100 runs.}
    \begin{tabular}{|l|l|c|c|c|c|c|c|c|c|}
    \hline
    \diagbox{$e$}{$\ell$}& Model & $2^{11}$ & $2^{12}$ & $2^{13}$  & $2^{14}$ & $2^{15}$ & $2^{16}$ & $2^{17}$ \\
    \hline 
    \multirow{2}{*}{0.25} & S5 & 0.126 & 0.214 & 0.401 & 0.798  & 1.508 &  2.980  & 5.805\\
    \cline{2-9} &          RTF & -     & -     & -     & -      & -     & -       & - \\   
    \hline   
    \multirow{2}{*}{0.5} &  S5 & 0.200 & 0.336 & 0.730 & 1.396  & 2.737 &  5.364  & 10.384\\
    \cline{2-9} &          RTF & -     &  -    & -     & -      & -     & -  & - \\   
    \hline   
    \multirow{2}{*}{1}   & S5  & 0.328 & 0.672 & 1.248 & 2.582  & 4.906 &  9.232  & 18.184\\
    \cline{2-9} &         RTF  & 0.102 & 0.373 & 0.770 & 1.558  & 3.443 & 8.345   & 17.824\\  
    \hline   
    \multirow{2}{*}{2} &   S5  & 0.623 & 1.194 & 2.317 & 4.566  & 8.573 &  17.137 & 34.219\\
    \cline{2-9} &         RTF  & 0.104 & 0.385 & 0.793 & 1.610  & 3.543 & 8.547   & 18.220\\ 
    \hline  
    \multirow{2}{*}{4} &   S5  & 1.156 & 2.312 & 4.403 & 8.334  & 16.830 & 33.545 & 67.033\\
    \cline{2-9} &         RTF  & 0.104 & 0.372 & 0.777 & 1.564  & 3.451  & 8.372  & 17.895\\
    \hline 
    \multirow{2}{*}{8} &   S5  & 2.214 & 4.363 & 8.328 & 16.438 & 33.714 & 67.336 & 137.632\\
    \cline{2-9} &         RTF  & 0.102 & 0.372 & 0.777 & 1.577  & 3.473  & 8.402  & 17.963\\
    \hline
    \end{tabular}
    \label{table:latency_vals}
\end{table}
\subsection{Long Range Arena Benchmark} \label{app:lra}

\subsubsection{Model Architecture Details} \label{app:lra_model}

For fair comparisons with S4 \cite{s4} and S4D \cite{s4d}, we employed the same model backbone, block design, and architectural hyperparameters as employed by S4. Each model contains a linear encoder and decoder that projects the inputs and outputs to an appropriate channel dimension. Simply put, each layer is a combination of a SSM layer, an activation function (GELU \cite{gelu}), followed by an output linear projection layer, and another activation function (GLU \cite{glu}), with skip connections \cite{resnet} and normalization applied before each every SSM and linear layer. Each channel in a SSM layer comprises of a SISO SSM with the ability to share the the transition matrix $\sA$ across channels through the number of SSMs hyperparameter (Num. SSM). This sets the number of unique $\sA$ matrices (or rational function denominator) which are then equally dispersed across the channel dimensions.
Additional hyperparameter details are outlined in Tables \ref{table:hyperparameter_lra} and \ref{table:hyperparameter_lra_layer}. 

Experiments using S4 and S4D models used the \texttt{PyKeops} implementation, available in the official S4 github repository \cite{s4}. Fused FFTConv \cite{flashfftconv} algorithms were \textbf{not} used for the RTF implementation.

\begin{table*}
    \centering
    \setlength{\tabcolsep}{4pt}
    \caption{Table with the hyperparameters used for classification datasets. BN and LN refer to Batch Normalization and Layer Normalization.}
    \begin{tabular}{@{}l|cccccccccccc@{}}
    \toprule
     & Layers & Channels & SSM State Size & Num. SSM & Norm. & Batch Size & Epochs \\
    \midrule 
    ListOps     & 6 & 256 & 4 & 1 & BN &  32 & 50 \\
    Text        & 6 & 256 & 4 & 1 & BN &  16 & 32 \\
    Retrieval   & 6 & 256 & 4 & 1 & BN &  64 & 20 \\
    Image       & 6 & 512 & 64 & 1 & BN & 50 & 200 \\
    Pathfinder  & 6 & 256 & 64 & 256 & BN & 64 & 200  \\
    Path-X      & 6 & 256 & 64 & 256 & BN & 16 & 50  \\
    \bottomrule 
    \end{tabular}
    \label{table:hyperparameter_lra}
\end{table*}

\begin{table*}
    \centering
    \setlength{\tabcolsep}{4pt}
    \caption{Table with the layer hyperparameters used for classification datasets.}
    \begin{tabular}{@{}lc|ccccc@{}}
    \toprule
     & Model & Dropout & LR & WD & SSM LR & SSM WD\\ 
    \midrule 
    \multirow{3}{*}{ListOps}    & S4 & 0.0 & 0.01 & 0.05 & 0.001 & 0.0\\
                                & S4D & 0.0 & 0.01 & 0.05 & 0.001 & 0.0\\
                                & RTF & 0.0 & 0.002 & 0.07 & - & -\\
    \midrule
    \multirow{3}{*}{Text}    & S4 & 0.0 & 0.01 & 0.05 & 0.001 & 0.0\\
                             & S4D & 0.0 & 0.01 & 0.05 & 0.001 & 0.0\\
                             & RTF & 0.1 & 0.005 & 0.05 & 0.0001 & 0.025 \\
    \midrule
    \multirow{3}{*}{Retrieval}    & S4 & 0.0 & 0.01 & 0.05 & 0.001 & 0.0\\
                                  & S4D & 0.0 & 0.01 & 0.05 & 0.001 & 0.0\\
                                  & RTF & 0.0 & 0.002 & 0.0 & 1e-6 & 0.0\\
    \midrule
    \multirow{3}{*}{Image}    & S4 & 0.1 & 0.01 & 0.05 & 0.001 & 0.0\\
                              & S4D & 0.1 & 0.01 & 0.05 & 0.001 & 0.0\\
                              & RTF & 0.1 & 0.006 & 0.05 & 0.005 & 0.05\\
    \midrule
    \multirow{3}{*}{Pathfinder}    & S4 & 0.0 & 0.004 & 0.05 & 0.001 & 0.0\\
                                   & S4D & 0.0 & 0.004 & 0.05 & 0.001 & 0.0\\
                                   & RTF & 0.1 & 0.002 & 0.05 & - & - \\
    \midrule
    \multirow{3}{*}{Path-X} & S4 & 0.0 & 0.002 & 0.05 & 0.001 & 0.0\\
                            & S4D & 0.0 & 0.002 & 0.05 & 0.001 & 0.0\\
                            & RTF & 0.1 & 0.001 & 0.05 & 0.001 & 0.0\\
    \bottomrule 
    \end{tabular}
    \label{table:hyperparameter_lra_layer}
\end{table*}

\subsubsection{Long Range Arena Benchmark Details}

The long range arena (LRA) benchmark \cite{lra} features 6 unique tasks within lengths of 1K-16K steps. These tasks involve diverse modalities and objectives, pushing models to reason about similarity, structure, and visuospatial relationships.

We offer additional context and specifics for each dataset from the LRA \cite{lra} that we examine, following the identical data pre-processing procedures as those used by \cite{s4}.

\begin{itemize}
\item \texttt{ListOps} An extended dataset introduced by \cite{listops}.  This task involves calculating the integer outcome of mathematical expressions encoded in prefix notation with brackets. Nested operations (min, max, etc.) and operands (0-9) are represented as one-hot vectors (17 unique values, brackets and operators combined). Sequence lengths vary, with max length of 2048. The dataset contains 10 distinct classes, each representing a possible integer outcome, with 96,000 training, 2,000 validation, and 2,000 test sequences.

\item \texttt{IMDB} Sentiment dataset from \cite{imdb}. This task involves classifying movie reviews into positive or negative sentiment categories based on sequences of integer tokens (encoded as one-hot vectors, 129 unique values). Sequence length varies, with a maximum length of 4,096. The dataset consists of 25,000 training and 25,000 test examples.

\item \texttt{Retrieval} This is derived from the ACL Anthology network corpus introduced by \cite{retrieval}. The datasets requires determining if two provided textual citations, encoded as a sequence of integer tokens, are the same. Characters are converted into a one-hot vector with 97 unique values. The two paired sequences can have different lengths, with a maximum sequence length of 4,000. There are two categories, signifying whether the citations are equivalent or not. The dataset comprises 147,086 training pairs, 18,090 validation pairs, and 17,437 test pairs.

\item \texttt{Image} The task utilizes the CIFAR-10 dataset introduced by \cite{cifar10}. It involves classifying a 32 × 32 grayscale CIFAR-10 image, presented as a one-dimensional raster scan, into one of ten categories. All sequences have the same length (1,024). The dataset comprises 45,000 training examples, 5,000 validation examples, and 10,000 test examples.

\item \texttt{Pathfinder} This is derived from the Pathfinder challenge, as presented by \cite{pathfinder}. It involves a 32 × 32 grayscale image that displays a start and an end point, each represented by a small circle. The image contains several dashed lines. The objective is to determine whether a dashed line (or path) connects the start and end points. There are two classes, signifying whether a valid path exists or not. All sequences have the same length (1,024). The dataset includes 160,000 training examples, 20,000 validation examples, and 20,000 test examples.

\item \texttt{Path-X} This is a variant of the Pathfinder challenge. With a longer sequence and more complex, in this version, the images are 128 × 128 pixels, leading to sequences that are sixteen times longer.

\end{itemize}

\subsection{Synthetic Memorization Tasks} \label{app:synthetic}

Both implementations of \textit{Copying} \cite{urnn} and \textit{Delay} \cite{hippo} were taken directly from the official S4 repository \cite{s4}, and was modified to enable drop in replacements of our RTF SSMs under identical conditions.

\subsubsection{\textit{Copying} Task}

Each model is first fed a $\ell_{\sf mem}$ length sequence of integer tokens randomly sampled from ${0,...,d-2}$, and then fed a $\ell_{\sf mem}$ length sequence of token number $d-1$ to recall the initial sequence. Table \ref{table:copying_hparams} lists the task hyperameters.
\begin{table}[h]
    \centering
    \setlength{\tabcolsep}{4pt}
    \caption{\textit{Copying} Task Hyperparameters.}
    \begin{tabular}{@{}l|c@{}}    
    \toprule
    Configuration & Value \\
    \midrule
    $\ell_{\sf mem}$ & 1024 \\
    Vocab. size $d$ & 64 \\
    Train-set Size & 10000 \\
    Test-set Size & 1000\\
    Batch Size & 8\\
    Epochs & 50\\
    LR & 0.001\\
    WD & 0.0\\
    \bottomrule 
    \end{tabular}
    \label{table:copying_hparams}
\end{table}

The overall model architecture is identical to that described in Section \ref{app:lra_model}. Each model was trained with 4 layers, 1024 channel dimensions, and the number of SSM was set to 1 (for weight sharing). Additionally, we initialized the RTF parameters by uniformly sampling from a range of 0 to 1, and applying the Montel constraint to limit the poles to a stable location.

\subsubsection{\textit{Delay} Task}

The models are given a signal of length $\ell_{\sf seq}$ and are tasked to output the original signal shifted by $\ell_{\sf delay}$ timesteps. The input is a white noise signal bandlimited to 1000 Hz. A single layer SSM with channel dimensions of 4 without a non-linear activation function was used for this experiment. Table \ref{table:delay_hparams} lists the task hyperameters.
\begin{table}[h]
    \centering
    \setlength{\tabcolsep}{4pt}
    \caption{\textit{Delay} Task Hyperparameters.}
    \begin{tabular}{@{}l|c@{}}    
    \toprule
    Configuration & Value \\
    \midrule
    $\ell_{\sf seq}$ & 4000 \\
    $\ell_{\sf delay}$ & 1000 \\
    Batch Size & 64\\
    Epochs & 20\\
    LR & 0.001\\
    WD & 0.0\\
    \bottomrule 
    \end{tabular}
    \label{table:delay_hparams}
\end{table}

%
%
%
%
%

\subsection{Laughing Hyena Distillation Task}

\begin{itemize}
    \item The baseline 160M parameter MultiHyena-Attention hybrid model consists of 6 Attention layers and 6 MultiHyena layers. 
    \item The distillation task aims to replace the Hyena filters in the 6 MultiHyena layers with an RTF or a modal SSM.
    \item Each MultiHyena layer consist of 256 independent SISO convolutional filters, which are projected to 768 dimensions as described in \cite{lhyena}.
    \item Both LH and RTF were trained for $1\mathrm{e}6$ iterations, on the AdamW \cite{adamw} optimizer with learning rates set to $1\mathrm{e}{-4}$. 
\end{itemize}

\subsection{WikiText103 Language Modeling} 

\subsubsection{Pilot Experiments} \label{app:wt-pilot}
We additionally compared S4, S4D, and RTF on WikiText103 under the modified Transformer backbone \cite{transformer-s4}, from the official S4 repository \cite{s4}, via drop-in replacements of S4 with S4D and RTF, while keeping the original hyperparameters. Table \ref{table:wt103-small} shows perplexity scores for the models across multiple state-sizes, trained for 25 epochs on two 40GB A100 GPUs.
The results show a consistent trend of RTF outperforming S4 and S4D across multiple state-sizes.

\begin{table}
\parbox{.45\linewidth}{
\centering
    \caption{
    WikiText103 language modeling perplexity scores (25 epochs).
   }
    \begin{tabular}{@{}l|cc@{}}
    \toprule
    Model & Perplexity $\downarrow$ \\
    \midrule 
    S4-4 &  26.86 \\
    S4D-4 & 26.98 \\
    RTF-4 & \textbf{26.36} \\
    \midrule 
    S4-64  & 26.82 \\
    S4D-64 & 26.67 \\
    RTF-64 & \textbf{26.01} \\
    \midrule 
    S4-256 &  - \\
    S4D-256 & 26.70 \\
    RTF-256 & \textbf{26.32} \\
    \bottomrule    
    \end{tabular}
    \label{table:wt103-small}
}
\hfill
\parbox{.45\linewidth}{
\centering
    \caption{Wikitext103 Hyperparameters.}
    \begin{tabular}{@{}l|c@{}}  
    \toprule
    Configuration & Value \\
    \midrule
    Sequence length & 1024 \\
    Batch Size & 16 (128 global)\\
    Epochs & 100\\
    LR & 0.001\\
    WD & 0.25\\
    Dropout & 0.25\\
    \midrule
    SSM State Size & 64 \\
    Channels & 768 \\
    Layers & 12 \\
    Low-Rank Dims. & 384 \\
    \bottomrule 
    \end{tabular}
    \label{table:wikitext_hparams}
}
\end{table}

\subsubsection{Model Architecture Details}

For our main WikiText103 experiment, we constructed Hyena-RTF by simply replacing the Hyena Filters in the Hyena Hierarchy model \cite{hyena} implemented in the \texttt{HazyResearch/safari} Github repository, with our RTF SSM. We also made slight modifications to the Hyena operator's output linear projection, by inserting an additional low-rank linear layer and a GELU \cite{gelu} activation, before the final output linear projection. This is to functionally mimic the low-rank MIMO SSM + non-linear activation function that Hyena-S5 \cite{s5} employs. It is worth noting that the additional low-rank layer does not increase parameter count since the original output linear projection also loses rank for compatibility of dimensions.
We observed that the zero-initialization alone was not enough for the model to stay within the stable region across training -- an important property for extrapolative tasks such as language generation. Therefore, we instead adopt the Xavier initialization \cite{glorotinit} over the rational function coefficients and apply the Montel constraint via an $\ell1$ penalization as shown in Section \ref{algo:l1_montel}. Table \ref{table:wikitext_hparams} lists the hyperparameters used to train our Hyena-RTF model.
\end{document}